\documentclass[times, review, 10pt]{elsarticle}

\usepackage{amsmath,amsfonts}
\usepackage{graphicx}
\usepackage{booktabs}
\usepackage{multirow}
\usepackage{url}
\usepackage{caption}
\usepackage[hidelinks]{hyperref}

\journal{Pattern Recognition}

\begin{document}

\begin{frontmatter}
\title{EGiME: Entropy-Guided Model Evaluation in Few-Shot Learning\tnoteref{t1}}

\author[gxu1]{Dequan Jin}
\ead{dqjin@gxu.edu.cn}
\author[gxu1]{Tao Li}
\ead{2406301025@st.gxu.edu.cn}
\author[gxu1]{Lingfeng Xing}
\ead{2406301052@st.gxu.edu.cn}
\author[shu1]{Shihui Ying}
\ead{shying@shu.edu.cn}
\author[gxu2]{Hui Zhao\corref{cor1}}
\ead{zhaohui\_2020@gxu.edu.cn}

\affiliation[gxu1]{organization={Department of Mathematics and Applied Mathematics, School of Mathematics and Information Science, Guangxi University}, city={Nanning}, state={Guangxi}, postcode={530004}, country={China}}
\affiliation[shu1]{organization={Department of Mathematics, School of Science, Shanghai University}, city={Shanghai}, postcode={200444}, country={China}}
\affiliation[gxu2]{organization={School of Economics, Guangxi University}, city={Nanning}, state={Guangxi}, postcode={530004}, country={China}}

\cortext[cor1]{Corresponding author}
\tnotetext[t1]{Abbreviations: EGiME, Entropy-Guided Model Evaluation; ViT, Vision Transformer; KNN, $K$-nearest neighbor; SVM, support vector machine; LEEP, Log Expected Empirical Prediction; LogME, Logarithm of Maximum Evidence; PACTran, PAC-Bayesian Transferability.}

\begin{abstract}
The feature-extraction quality of pre-trained models sets the ceiling of downstream transfer learning, yet few-shot model selection remains difficult: supervised evaluation on scarce labels is unstable and yields fluctuating rankings. We propose Entropy-Guided Model Evaluation (EGiME), a semi-supervised criterion that assesses a candidate model from the clustering structure of its feature space without any classifier training. EGiME performs $K$-means clustering over all labeled and unlabeled samples, then combines two entropy terms: an inter-cluster entropy measuring global separability, and an intra-cluster entropy measuring the compactness of labeled anchors, fused through an inverse-frequency weight that emphasizes small clusters. On five few-shot benchmarks and nine pre-trained models, EGiME achieves a correlation score of 0.966 with the full-pool ground-truth accuracy obtained by averaging the accuracies of K-means, KNN, and SVM. At labeling ratios of 4\%--20\%, it produces markedly more stable model rankings than cross-validated supervised baselines, establishing it as a practical alternative to conventional supervised evaluation for pre-trained model selection under data scarcity.
\end{abstract}


\end{frontmatter}

\section{Introduction}

The powerful feature extraction capability of deep neural networks is fundamental in image recognition tasks. It enables the mapping of raw pixel data into high-dimensional representations with distinct semantic discriminability, determines the generalization, transferability, and robustness of network models~\cite{he2019rethinking,dosovitskiy2021vit}, and serves as a core prerequisite for stable task performance in challenging scenarios such as few-shot learning and domain adaptation. In practical industrial and academic scenarios including industrial defect detection, medical image analysis, and fine-grained species identification, annotated sample scarcity is a ubiquitous and restrictive constraint. Accordingly, the capability validation and quantitative evaluation of deep models in few-shot settings have become a key bottleneck limiting the practical deployment of deep learning models. Therefore, exploring stable, interpretable quantitative evaluation methods for feature extraction capabilities in few-shot scenarios is of significant theoretical research value and practical engineering significance.

Existing evaluation paradigms for the feature extraction capacity of deep neural networks fall into two mainstream categories. The first way corresponds to indirect evaluation grounded on downstream task performance, which adopts task metrics such as classification accuracy and Top-k precision to quantify feature quality, serving as a standard protocol across academia and industry. Nevertheless, such methods heavily rely on large-scale annotated test sets, and their outcomes are susceptible to irrelevant confounding factors including classifier design and training strategies. In few-shot learning tasks, these evaluation methods suffer severe fluctuations due to prominent sampling bias, failing to eliminate interferences introduced by task layers and reflect the intrinsic model performance in feature extraction. The second way to directly evaluate model performance is based on geometric properties of the feature space. It characterizes the distribution property of ``intra-class compactness and inter-class separability'' via metrics such as intra-cluster and inter-class distances and silhouette coefficients, delivering superior interpretability. However, these approaches compute metrics using labeled samples and yield poor statistical stability in few-shot learning tasks. Furthermore, they fail to exploit global distribution constraints embedded in massive unlabeled samples, and thus cannot comprehensively measure the overall quality of the feature space.

To address these issues, cross-validation acts as a widely adopted statistical augmentation technique. It iteratively partitions and cross-reuses limited labeled samples. Its representative implementations include k-fold cross-validation and leave-one-out cross-validation. Cross-validation reduces evaluation variance induced by single random splits, improving the confidence and stability of generalization performance. Within the N-way K-shot evaluation protocol in few-shot learning, the standard pipeline that repeatedly samples support and query sets and averages the corresponding accuracy metrics essentially constitutes a task-specific variant derived from the cross-validation paradigm. Nevertheless, cross-validation merely reuses available annotated samples without lifting the limitation of limited annotation volume, nor can it incorporate global distribution information from unlabeled samples to fundamentally mitigate systematic biases inherent to few-shot evaluation. Furthermore, cross-validation only outputs statistical distributions of task metrics and lacks interpretability at the feature space level, making it incapable of revealing intrinsic structural discrepancies in the feature extraction capacity of different models.

Semi-supervised representation learning offers data-oriented solutions to address the issues of feature evaluation with very few samples. Advances in semi-supervised learning and self-supervised contrastive learning have fully demonstrated that abundant unlabeled samples can impose global distribution constraints on the semantic structure of feature spaces. Consistency regularization and contrastive loss guide models to learn more generalizable feature representations. Their effectiveness has been extensively validated in tasks such as few-shot classification and cross-domain transfer learning. Nevertheless, most existing studies focus on ``how to leverage unlabeled samples to boost feature extraction performance'', while rarely reversing this logic to adopt distribution information of unlabeled samples for ``evaluating feature extraction capacity''. It prevents the full exploitation of available samples to evaluate model performance on feature extraction in few-shot learning scenarios.

From the perspective of measurement tools, information entropy, a classical metric for characterizing the disorder degree of data distributions, has been employed in uncertainty estimation, representation quality assessment, and feature selection for deep models. Several studies quantify representation discriminability by calculating entropy values of feature dimensions or category distributions. Nonetheless, most existing entropy-based evaluation metrics still rely on labeled samples for measurement construction, excluding the global distribution entropy derived from unlabeled samples in evaluation frameworks. Meanwhile, a unified entropy measurement framework oriented toward the core criterion of ``intra-class compactness and inter-class separability'' remains absent. This makes it infeasible to comprehensively quantify feature extraction capacity. In summary, current literature has not established a unified few-shot feature evaluation paradigm integrating semi-supervised data utilization and information entropy measurement, and the distribution value of unlabeled samples during evaluation procedures has not been systematically excavated.

According to the above discussion, we can see that when labeled samples have become extremely scarce, any evaluation paradigm that relies on training a classifier with labeled data is inherently unsustainable.  To address this issue, distribution information from unlabeled samples must be incorporated to reliably evaluate the representation quality of models. On the other hand, if the distribution of extracted samples in the feature space satisfies the core feature quality criterion of intra-class compactness and inter-class separability and exhibits a clear and well-defined clustering structure, the model can be regarded as possessing strong feature extraction capability. Motivated by this insight, this paper proposes a novel Entropy-Guided Model Evaluation (EGiME) method. EGiME first extracts feature representations of all available samples (both labeled and unlabeled) using each candidate pre-trained model, and then performs $K$-means clustering in the resulting feature space to reveal their latent geometric topology. Subsequently, entropy-guided metrics are formulated to quantitatively characterize the quality of the clustering configuration. Specifically, an inter-cluster entropy is defined to measure the uniformity of the sample distribution across different clusters, capturing global separability; concurrently, an intra-cluster entropy is introduced to evaluate the concentration of labeled samples within each individual cluster, reflecting local compactness. Finally, these two complementary indicators are fused into a unified metric that serves as a comprehensive measure of model performance. The primary contributions of this work are summarized as follows:
\begin{enumerate}
	\item We propose a novel semi-supervised evaluation paradigm for model assessment in few-shot scenarios. It breaks the purely supervised constraint of traditional pre-trained model evaluation and systematically integrates the global distribution of massive unlabeled samples into the feature extraction capability evaluation workflow, fundamentally alleviating the core issues of unstable model ranking caused by labeled data scarcity in few-shot scenarios.

	\item We provide a unified two-dimensional evaluation metric constructed on information entropy. Taking information entropy as the core theoretical instrument, we design the inter-cluster entropy and intra-cluster entropy to characterize the global separability of semantic clusters and the local compactness of homogeneous samples within each cluster and fuse them into a comprehensive metric, realizing accurate and interpretable quantification of the core feature quality principle of intra-class compactness and inter-class separability.

	\item We present a lightweight evaluation pipeline. The proposed method employs $K$-means clustering as the unsupervised structure discovery module. The entire evaluation procedure requires no auxiliary classifier training or fine-tuning of pre-trained models, which technically circumvents evaluation distortion induced by classifier overfitting under few-shot settings. With only the true class number of the target dataset as the prior parameter, this method features high computational efficiency and strong cross-domain adaptability, enabling efficient model selection for large-scale pre-trained model zoos.
\end{enumerate}

The remainder of this paper is organized as follows. Section~2 reviews related work on transferable model selection, feature selection, and entropy-based clustering analysis. Section~3 presents the methodology. Section~4 reports the experimental setup and results. Section~5 concludes the paper and outlines future research directions.

\section{Related Work}

\subsection{Model Selection in Transfer Learning}

Selecting an appropriate feature extractor for a downstream task is a critical step that determines ultimate performance. With the proliferation of pre-trained models such as AlexNet~\cite{krizhevsky2012alexnet}, VGG~\cite{simonyan2015vgg}, GoogLeNet~\cite{szegedy2015googlenet}, ResNet~\cite{he2016resnet}, and Vision Transformers (ViT)~\cite{dosovitskiy2021vit} in public model zoos, efficiently choosing the best model for a target task has become a fundamental challenge, formally defined as the model selection problem~\cite{zamir2018taskonomy}.

The importance of model selection lies in the observation that different backbone architectures exhibit significantly varying representation capabilities on the same target dataset, and selecting an unsuitable backbone leads to degraded downstream performance. To avoid the costly full fine-tuning of every candidate model, researchers have proposed the research direction of transferability estimation~\cite{you2021logme}, which aims to design a metric that can predict the performance of a pre-trained model on a target dataset without exhaustive fine-tuning. Representative works include the LogME method based on Bayesian evidence maximization~\cite{you2021logme}, the LEEP method based on empirical conditional probability~\cite{nguyen2020leep}, the H-score method from information theory~\cite{bao2019hscore}, the ETran method based on energy models~\cite{gholami2023etran}, the PACTran method based on PAC-Bayes theory~\cite{ding2022pactran}, and the NCTI method based on neural collapse distance~\cite{wang2023ncti,papyan2020neuralcollapse}, which evaluates transferability by measuring the deviation of the feature space from the ideal simplex structure predicted by the neural collapse phenomenon. Furthermore, systematic evaluations of multiple transferability metrics across different benchmark datasets have been conducted~\cite{agostinelli2022stable,ibrahim2023newer}, revealing the applicability differences of various methods in different scenarios.

The mainstream approach to evaluating candidate models is $k$-fold cross-validation~\cite{kohavi1995crossvalidation}. It divides the limited labeled samples into $k$ subsets, alternately trains auxiliary classifiers such as linear SVM or KNN, and averages the accuracy. This approach provides asymptotically unbiased estimates when labeled samples are sufficient.

Applied directly to few-shot scenarios, these methods become unreliable~\cite{perez2021truefewshot,vabalas2019cv}. The compressed per-fold training set cannot support a stable auxiliary classifier, inflating the variance of accuracy estimates and increasing sensitivity to individual support samples. Moreover, as a purely supervised paradigm, cross-validation discards the structural information in unlabeled data, which constitutes the vast majority in practice. These factors cause ranking fluctuations across sampling rounds and amplify overfitting when the sample count is far below the feature dimensionality.

The core of model selection is to discriminate the quality of features extracted by candidate models, so the problem above essentially reduces to feature-quality evaluation under few-shot conditions---which motivates us to examine how the feature selection community has addressed similar dilemmas.

\subsection{Feature Selection Methods in Few-Shot Scenarios}

Feature selection identifies the most discriminative subset or transformation from a high-dimensional feature space~\cite{hancer2020fssurvey}, a challenge especially acute in few-shot learning~\cite{vinyals2016matching,snell2017protonet,finn2017maml,wang2020fslsurvey}. Traditional methods fall into three categories: filter methods evaluate features independently through statistical indicators such as mutual information and correlation coefficients, offering efficiency but ignoring inter-feature interactions; wrapper methods embed selection into classifier training through recursive elimination or heuristic search, capturing interactions at high computational cost and overfitting risk; and embedded methods treat selection as a regularization constraint during training, such as L1 regularization in Lasso.

In few-shot scenarios these methods face the $p \gg n$ dilemma: feature-label correlation measures are highly susceptible to sampling noise, and wrapper and embedded methods overfit with insufficient samples, yielding selected features that generalize poorly. Recent few-shot-oriented variants design automatic selection modules that weight features from different backbones, or combine mid-level transferability with high-level discriminability for cross-domain tasks; however, most remain embedded in the training pipeline of a specific model and cannot be directly transferred to cross-model comparison.

Clustering structures~\cite{caron2018deepcluster} and information entropy~\cite{jin2021weighting} have long served as effective tools for measuring feature quality: information compression indices evaluate feature importance through clustering changes after feature removal, mutual information measures feature-label correlation, and clustering purity and silhouette coefficients reflect inter-class separation. The core insight is that the clustering structure reflects feature discriminability, and entropy provides a natural tool for quantifying it. This motivates our study: taking the clustering structure as the analysis object and information entropy as the quantification tool, we construct a feature-quality criterion free of classifier training that addresses both scarce labels and cross-model comparison.

\subsection{Feature Space Distribution Analysis Methods}

Few-shot feature-quality evaluation requires a structure-discovery tool that exploits unlabeled data without supervision. Self-supervised and vision-language pre-training have shown that unlabeled data alone can yield highly transferable representations~\cite{chen2020simclr,he2020moco,he2022mae,radford2021clip,oquab2024dinov2}, while supervised contrastive learning further improves the learned embedding when labels are available~\cite{khosla2020supcon}. Earlier semi-supervised learning established the same value of unlabeled structure~\cite{chapelle2006ssl}, with a frozen embedding already competitive in few-shot classification~\cite{tian2020goodembed}. Cluster analysis provides precisely such a tool~\cite{macqueen1967kmeans}, and variants such as nonlinear manifold clustering~\cite{abdolali2021manifold} and self-supervised clustering~\cite{caron2020swav} enrich the toolbox.

Cluster analysis partitions data so that intra-cluster similarity is maximized and inter-cluster similarity is minimized. Requiring no labels, it naturally suits few-shot scenarios by using all samples, including the unlabeled majority, to build a global geometric description of the feature space. Since the true number of target categories is known, the cluster number is set to $K=s$ without additional tuning. We adopt classic $K$-means~\cite{macqueen1967kmeans}, whose fixed-centered variant stabilizes centers against noisy initializations~\cite{ay2023fckmeans}. $K$-means is widely adopted for its intuitive principle, simple implementation, and computational efficiency. In EGiME, we avoid relying on random restarts: for each labeled fold, the class-mean feature vector of every class is used as a centroid, and $K$-means is run once from this initialization ($n_{\mathrm{init}}=1$) over the complete feature set. This class-mean initialization anchors the clusters to the available semantic information and reduces the variability that would arise from a single random initialization. Its workflow takes all sample features as input, searches for $K$ centers, assigns each sample to the nearest center, and outputs cluster assignments reflecting the intrinsic topology of the feature space.

Cluster assignments are converted into model-quality signals through information entropy~\cite{shannon1948,cover2006elements}: inter-cluster entropy measures the uniformity of sample distribution across clusters to reflect global separability, while intra-cluster entropy measures the concentration of labeled samples within each cluster to reflect local compactness. Combining the two with an inverse frequency weighting yields the unified weighted entropy metric, detailed in Section~3.

\section{Methodology}

\subsection{Motivation}

If a model can extract sample features effectively, the clusters formed by samples in the feature space should be consistent with their semantic categories. That is, samples of the same category form a single cluster, and clusters corresponding to different categories are well separated. Nevertheless, practical models often fail to satisfy this ideal property. It is possible that the extracted features of two samples lie close to each other in the feature space although they belong to distinct categories. Meanwhile, due to considerable intra-class variance, samples from the same category may scatter across multiple clusters.

\begin{figure}[!htb]
\centering
\begin{minipage}[b]{0.49\linewidth}
\centering
\includegraphics[width=\linewidth]{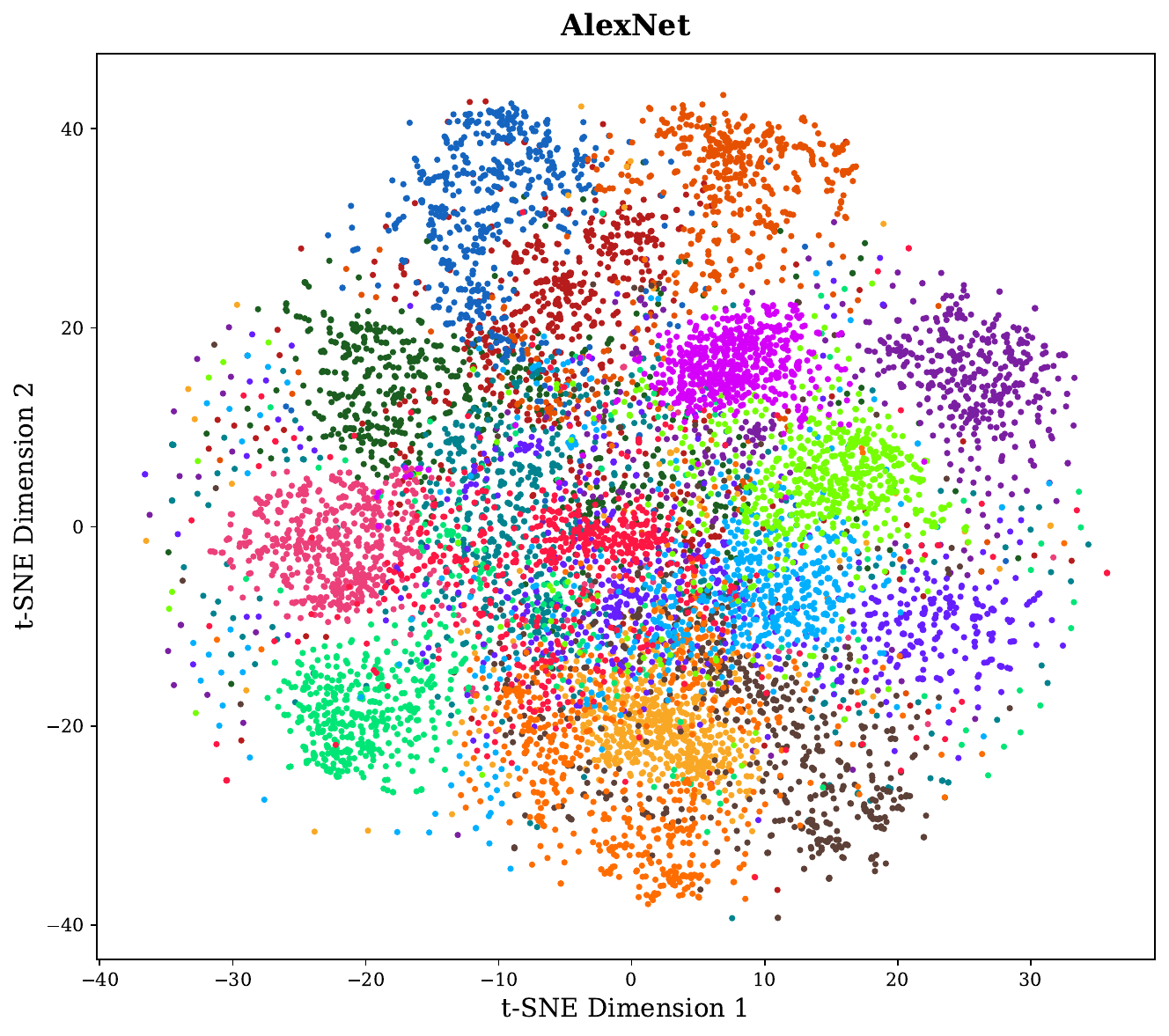}
\centerline{(a) AlexNet}
\end{minipage}
\hfill
\begin{minipage}[b]{0.49\linewidth}
\centering
\includegraphics[width=\linewidth]{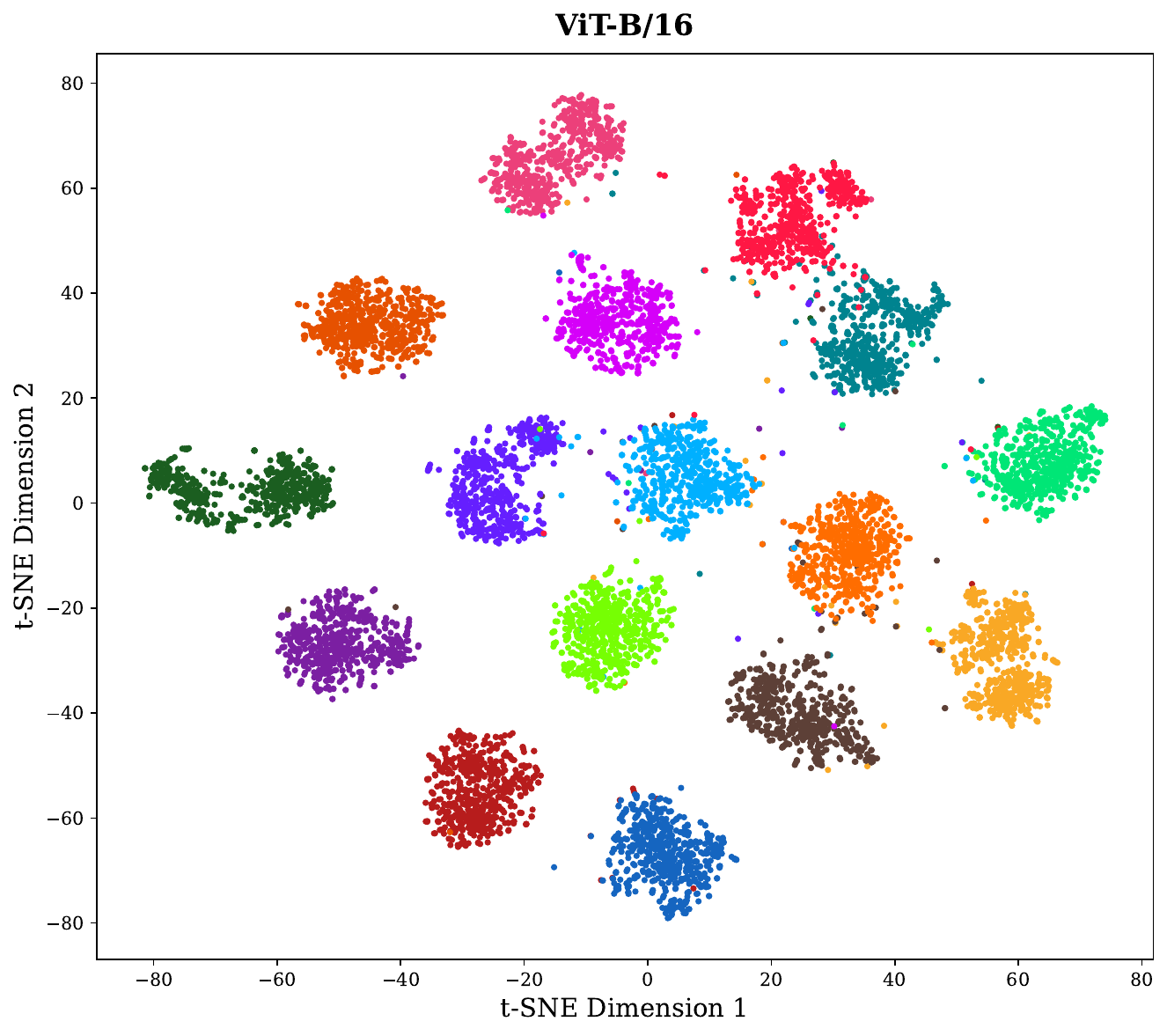}
\centerline{(b) ViT-B/16}
\end{minipage}
\caption{ The feature-extracted sample distributions derived by AlexNet (left) and ViT-B/16 (right) from the Mini-ImageNet dataset and visualized by $t$-SNE. Different colors represent different categories.}
\label{fig:tsne}
\end{figure}

This phenomenon can be intuitively verified from the visualization results in Fig.~\ref{fig:tsne}, which shows the $t$-SNE~\citep{van2008tsne} projection of feature distributions of AlexNet (left) and ViT-B/16 (right) on the Mini-ImageNet dataset. It can be clearly observed that for the high-performance ViT-B/16 model, samples of the same class are tightly clustered in the feature space, with clear boundaries between different classes, and the clustering structure exhibits a distribution pattern highly consistent with true semantic categories. For AlexNet, the feature distribution is relatively chaotic, with samples of the same class dispersed and severe overlap across classes, making them difficult to distinguish. This observation indicates that the degree of consistency between the clustering structure and the true categories is an effective indicator of feature extractor quality: the higher the consistency, the stronger the model's representational capability. Therefore, if we have a way to measure this consistency, we can evaluate the model's representational capability.

However, under few-shot settings, the scarcity of labeled samples impedes accurate quantification of the consistency between sample clustering structures and ground-truth categories. To circumvent this obstacle, an alternative evaluation strategy is required. As illustrated in Fig.~\ref{fig:tsne}, compared with AlexNet, ViT with superior feature extraction performance generates more uniformly distributed clusters within the feature space. Meanwhile, each cluster almost exclusively contains samples belonging to a single category. This observation inspires us to adopt these two properties as surrogate criteria for measuring the consistency between clustering structures and true categories. Specifically: when all samples (including unlabeled ones) form more evenly distributed clusters in the feature space, and the labeled samples within each cluster are purer, the corresponding model possesses stronger feature extraction capability. Given the above definition, the subsequent question becomes how to measure the distribution uniformity of features within the feature space.

In information theory, entropy is a widely adopted metric for quantifying information volume. For a discrete distribution over $K$ groups, let $p_j=n_j/N$, where $n_j$ is the number of samples in the $j$-th group and $N$ is the total number of samples. The Shannon entropy and its normalized form are $H=-\sum_{j=1}^{K}p_j\log p_j$ and $E=H/\log K$, respectively. Thus, $E\in[0,1]$, with $E=0$ when all samples fall into one group and $E=1$ when they are uniformly distributed across the $K$ groups. Entropy decreases when samples are distributed more compactly and increases as the distribution becomes more uniform, so it can measure the uniformity of sample distribution.

To calculate information entropy, it is necessary to acquire the distribution frequency of samples in the feature space. To this end, the feature space first needs to be discretized into multiple grid cells, and the frequency is then calculated based on these grids. Nevertheless, as illustrated in Figure 1, the grid-based frequency computation breaks the integrity of sample clusters, where a single cluster may be split across multiple grids. In addition, the choice of grid size exerts a substantial impact on the estimated sample frequency distribution. Under few-shot settings, it is difficult to determine an appropriate grid partition relying on sample labels. A more natural alternative is to compute distribution frequencies according to the sample clusters formed in the feature space, which requires clustering algorithms to capture the spatial distribution patterns of samples. As an unsupervised approach, clustering does not depend on sample labels. Hence, it can effectively characterize the spatial aggregation of samples even under few-shot constraints and eliminate disturbances to frequency calculation arising from different grid partition strategies. On this basis, we need to construct two entropy metrics to characterize the properties of sample distribution: one measures the macroscopic distribution uniformity of sample clusters, and the other quantifies the purity of labeled samples within each cluster.

\begin{figure}[!htb]
\centering
\includegraphics[width=0.8\textwidth]{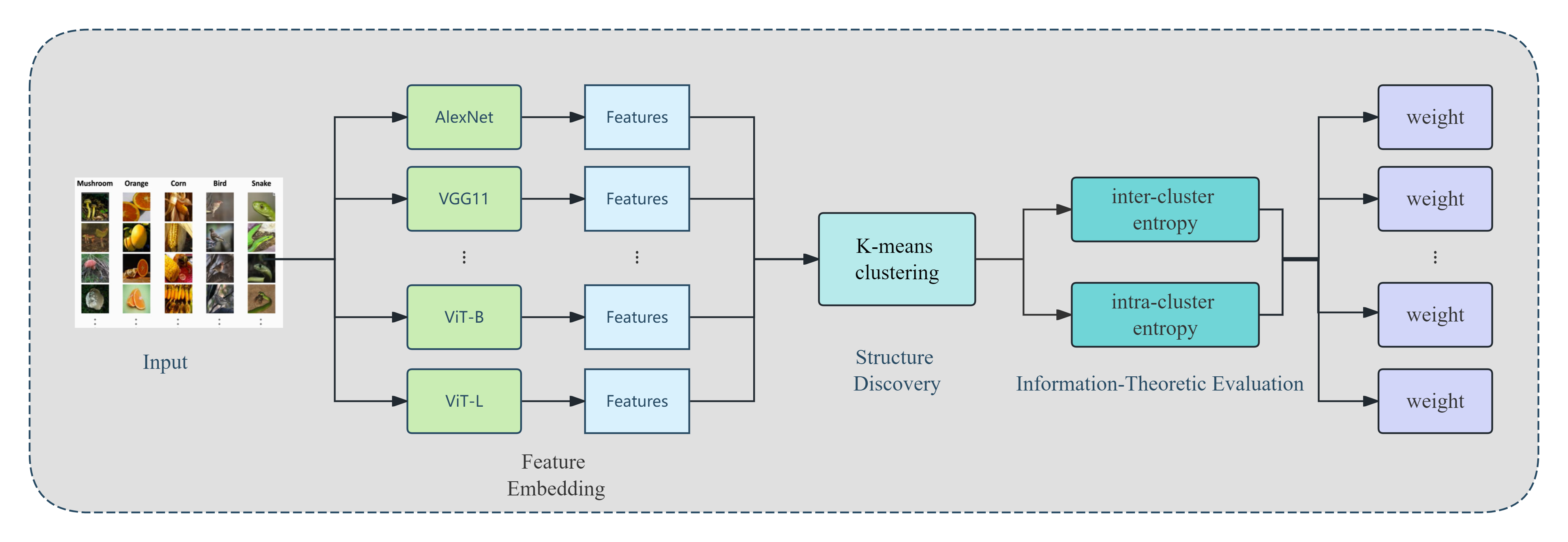}
\caption{Overall framework of the proposed EGiME methodology. The pipeline consists of three main stages: Feature embedding, structure discovery, and entropy-based evaluation.}
\label{fig:framework}
\end{figure}

Motivated by the above discussion, we propose an Entropy-Guided Model Evaluation (EGiME) method to infer the candidate models' capability on feature extraction by clustering all samples in the corresponding feature spaces and examining their consistency between sample distributions and semantic categories with limited labels. As shown in Fig.~\ref{fig:framework}, the EGiME pipeline consists of three main stages. First, candidate pre-trained models map all input images to high-dimensional feature vectors, completing feature embedding. Second, K-means clustering is performed in the feature space, with the number of clusters naturally set to the number of target categories. Third, based on the clustering results, the labeled samples serve as semantic references, and entropy-based metrics simultaneously quantify the inter-class separability and intra-class compactness of the feature space, ultimately providing a comprehensive evaluation score of model performance. The following sections will elaborate on the implementation details of these three stages.

\subsection{Problem Definition and Notation}

Let $D = \{y_1, y_2, \dots, y_n\}$ be an dataset, where $y_i$ represents the raw image. Denote the set of $q$ pre-trained models $\mathcal{M} = \{M_1, M_2, \dots, M_q\}$. For any candidate model $M_r \in \mathcal{M}$, it is regarded as a feature extraction function $f_r: \mathcal{Y} \rightarrow \mathbb{R}^m$, which maps a raw image into an $m$-dimensional feature vector. Consequently, we obtain the corresponding feature set $D_f^{(r)}$, where each sample corresponds to a feature vector expressed as $x_i^{(r)} = f_r(y_i)$.

Among the total $n$ samples, only $l$ samples are provided with category labels, constituting the label set $D_l$. These labels belong to the category set $\mathcal{L}$ with cardinality $|\mathcal{L}| = n_l$. The total number of true categories in the entire dataset is denoted as $s$ (typically $s \ge n_l$), which is assumed to be known as prior information. In the following discussion, we suppose $s = n_l$.

Our objective is to construct a quantitative metric $E_r$ based on the feature set $D_f^{(r)}$ and the labeled set $D_l$, which evaluates the feature extraction capability of the model $M_r$ on the target domain. Furthermore, this metric must maintain high stability and discriminability even in few-shot scenarios where labeled samples are extremely scarce.

\subsection{Feature Space Clustering and Inverse Frequency}

To evaluate the structural characteristics of the feature space, K-means clustering is performed on the feature space $D^{(r)}$, where the number of clusters is set to $K =s$. This yields the cluster assignment vector $\mathbf{c}^{(r)}$, in which $c_i^{(r)}$ denotes the cluster to which the $i$-th sample belongs.

The cluster count vector is defined as $\mathbf{Cr} = (cr_1, cr_2, \dots, cr_K)^\mathrm{T}$, where $cr_j$ is the number of samples in the $j$-th cluster.

Subsequently, the raw cluster frequency vector is defined as $\mathbf{f}_r = (f_{r1}, f_{r2}, \dots, f_{rK})^\mathrm{T}$ with $f_{rj} = cr_j / n$.

To enhance the sensitivity to small clusters, an inverse frequency weight is proposed: $\tilde{f}_{rj} = \frac{1 / f_{rj}}{\sum_{k=1}^K (1 / f_{rk})}$ for $j = 1, \dots, K$.
This weighting strategy explicitly assigns higher significance to low-frequency clusters (i.e., small clusters), thereby highlighting the model's discriminative capability for hard samples and rare categories during the evaluation process.

\subsection{Inter-cluster Entropy and Intra-cluster Entropy}

The inter-cluster entropy measures the uniformity of the sample distribution across the $K$ clusters:
\begin{equation}
E_r^{(\mathrm{inter})} = -\frac{1}{\log K} \sum_{j=1}^K \tilde{f}_{rj} \log \tilde{f}_{rj}
\label{eq:inter_entropy}
\end{equation}
A larger value indicates that the samples are uniformly dispersed across the clusters, reflecting superior global separation capability.

For labeled samples, we consider the distribution of each category within the clusters. Let $\mathcal{I}_i = \{ j \in \{1,\dots,l\} : l_j = i \}$ denote the index set of labeled samples for category $i$ with $N_i = |\mathcal{I}_i|$, and let the class-cluster count vector be $\mathbf{n}_{ri} = (n_{ri1}, n_{ri2}, \dots, n_{riK})^T$ with $n_{rij} = \sum_{k \in \mathcal{I}_i} \mathbb{I}[c_k^{(r)} = j]$.

Since the labeled samples are evenly distributed among classes in most few-shot scenarios, we form the frequency vector $\mathbf{f}_{ri} = (f_{ri1}, f_{ri2}, \dots, f_{riK})^T$ without inverse weighting. The intra-cluster entropy is $E_{ri}^{(\mathrm{intra})} = -\frac{1}{\log K} \sum_{j=1}^K f_{rij} \log f_{rij}$, and the average intra-cluster entropy is $E_r^{(\mathrm{intra})} = \frac{1}{n_l} \sum_{i=1}^{n_l} E_{ri}^{(\mathrm{intra})}$. A smaller $E_r^{(\mathrm{intra})}$ indicates that samples of the same class are highly concentrated within clusters, demonstrating better local compactness.

\subsection{Entropy-guided Metric for Model Evaluation}
In the above discussion, a larger $E_r^{(\mathrm{inter})}$ indicates superior global separation while a smaller $E_r^{(\mathrm{intra})}$ demonstrates better local compactness. An excellent feature extractor should simultaneously maximize the former and minimize the latter, so we evaluate model performance by combining them as follows:
\begin{equation}
E_r = \frac{E_r^{(\mathrm{inter})}}{1 + E_r^{(\mathrm{intra})}}
\label{eq:weighted_entropy}
\end{equation}

This metric reflects the trade-off between local compactness and global separability, and a larger $E_r$ indicates superior model performance. The constant 1 in the denominator prevents division by zero. To reduce the dependence of the score on a single partition of the scarce labeled samples, we adopt a fold-averaged estimation procedure. Specifically, the labeled samples are divided into $F$ stratified folds, where $F=\min(10,\lfloor n_{\mathrm{shot}}/2\rfloor)$; thus, $F$ ranges from 2 to 10 for $n_{\mathrm{shot}}\in[4,20]$. For each fold, the class-mean feature vector of every class in the current labeled fold is used as the corresponding $K$-means centroid, and $K$-means is run once from this initialization ($n_{\mathrm{init}}=1$) on the complete evaluation pool containing both labeled and unlabeled samples. The resulting assignments are then used to compute the inter- and intra-cluster entropies, with all labeled samples serving as semantic references. The final score is the average of the fold-level weighted entropies, $\bar{E}_r=F^{-1}\sum_{j=1}^{F}E_{rj}$, where $E_{rj}$ denotes the weighted entropy obtained in fold $j$. Algorithm~1 summarizes this fold-averaged procedure.

\begin{table}[!htb]
\centering
\caption{EGiME: Fold-Averaged Entropy-Guided Model Evaluation}
\label{alg:weighted}
{\small
\begin{tabular}{p{0.95\linewidth}}
\toprule
\textbf{Algorithm 1} Fold-Averaged EGiME \\
\midrule
\textbf{Input:} Raw image dataset $\mathcal{D}$, model set $\mathcal{M} = \{M_1, \dots, M_q\}$, cluster number $K=s$, labeled sample indices and corresponding labels, number of stratified folds $F$ \\
\textbf{Output:} Fold-averaged weighted entropy vector $\bar{\mathbf{E}} = (\bar{E}_1, \dots, \bar{E}_q)^T$ \\
\midrule
\textbf{1: for} each model $M_r \in \mathcal{M}$ \textbf{do} \\
\textbf{2:} \; Extract the full-pool feature set $D_f^{(r)}$ and partition the labeled samples into $F$ stratified folds; \\
\textbf{3:} \; \textbf{for} each fold $j=1,\dots,F$ \textbf{do} \\
\textbf{4:} \; \quad Compute the class-mean centroids of the current labeled fold, and run K-means once from these centroids ($n_{\mathrm{init}}=1$) on the complete pool $D_f^{(r)}$; \\
\textbf{5:} \; \quad Compute the inter- and intra-cluster entropies and the fold-level score $E_{rj}=E_{rj}^{(\mathrm{inter})}/(1+E_{rj}^{(\mathrm{intra})})$; \\
\textbf{6:} \; \textbf{end for} \\
\textbf{7:} \; Compute $\bar{E}_r=F^{-1}\sum_{j=1}^{F}E_{rj}$; \\
\textbf{8: end for} \\
\textbf{9: return} $\bar{\mathbf{E}}$ \\
\bottomrule
\end{tabular}
}
\end{table}

\section{Experiments}

\subsection{Experimental Setup}

\textbf{Datasets:} We evaluate our proposed method on five widely-used few-shot learning benchmark datasets: the validation set of Mini-ImageNet (Mini-val, 16 classes), STL-10 (10 classes, $96 \times 96$ pixels), CIFAR-10 (10 classes, $32 \times 32$ pixels), EuroSAT (10 classes, $64 \times 64$ pixels), and CIFAR-100 (100 classes, $32 \times 32$ pixels).

\textbf{Evaluation Pool:} To construct a unified and controllable evaluation benchmark, we draw samples from the labeled data source used for each dataset. Specifically, 100 images are randomly selected from each class to form the evaluation pool. All selected samples participate in the subsequent feature-based clustering and evaluation.

\textbf{Experimental Setting:} $n_{\mathrm{shot}}$ samples are randomly selected per class as labeled samples from the evaluation pool, where $n_{\mathrm{shot}} \in \{4, 6, 8, 10, 12, 14, 16, 18, 20\}$, simulating different levels of few-shot labeling. The remaining $100-n_{\mathrm{shot}}$ samples serve as unlabeled data. Such sampling ensures that the proportion of labeled samples ranges from 4\% to 20\%, effectively simulating real-world scenarios with extremely scarce labeled samples. Cross-validation (CV) is adopted for comparison: standard 10-fold cross-validation when $n_{\mathrm{shot}} \ge 10$, and $n_{\mathrm{shot}}$-fold cross-validation (i.e., leave-one-out) when $n_{\mathrm{shot}} < 10$, ensuring that each fold contains at least one representative sample from each class, avoiding evaluation bias caused by the absence of certain classes in the validation set due to insufficient samples. For each $n_{\mathrm{shot}}$, the random sampling process is repeated 100 times to eliminate statistical variance.

\textbf{Models:} We evaluate nine backbone architectures pre-trained on the ImageNet dataset, covering traditional convolutional neural networks (AlexNet, VGG11, VGG13, GoogLeNet, ResNet18, ResNet34) and modern Vision Transformers (ViT-S, ViT-B/16, ViT-L). The detailed specifications of these models, including their architecture types and extracted feature dimensions, are summarized in Table~\ref{tab:models}. Since the original classification heads of different network models yield inconsistent impacts on the network performance, we discard them and retain only the feature extraction backbone of each network. The resulting features are used to compute KNN, SVM, and K-means accuracies solely for constructing the full-label reference ranking described below; EGiME itself does not train a classifier. For CNN-based models, feature vectors are extracted directly after the global average pooling layer; for ViT models, the output vector corresponding to the \texttt{[CLS]} token is extracted as the image representation.

\begin{table}[!htb]
\centering
\caption{Specifications of the Pre-trained Models Evaluated in the Experiments}
\label{tab:models}
\resizebox{0.6\textwidth}{!}{%
\begin{tabular}{llc}
\toprule
\textbf{Model} & \textbf{Architecture} & \textbf{Feature Dim.} \\
\midrule
AlexNet   & CNN                & 4096 \\
VGG11     & CNN                & 4096 \\
VGG13     & CNN                & 4096 \\
GoogLeNet & CNN                & 1024 \\
ResNet18  & CNN                & 512  \\
ResNet34  & CNN                & 512  \\
ViT-S     & Vision Transformer & 384  \\
ViT-B/16  & Vision Transformer & 768  \\
ViT-L     & Vision Transformer & 1024 \\
\bottomrule
\end{tabular}%
}
\end{table}

\textbf{Ground-truth Ranking and Evaluation Metrics:} At present, there are no criteria for quantitatively evaluating the feature extraction performance of different deep neural network models. We therefore define the ground-truth score of a model as the equal-weight mean of the accuracies of three complementary predictors evaluated over the full evaluation pool: KNN with one neighbor, a linear SVM, and K-means initialized with class means so that the cluster indices align with the class labels. For each model, the mean accuracy of each predictor is computed independently by 10-fold cross-validation over the fully labeled evaluation pool. These three mean accuracies are then averaged with equal weights to obtain the ground-truth score, and the nine models are ranked by this score (rank 1 denotes the highest accuracy; ties receive the average rank). In other words, the average is taken over accuracies, not over model rankings. KNN and SVM capture local neighborhood consistency and supervised decision-boundary quality, whereas K-means supplements them with the global geometric organization of the feature space, rewarding intra-class compactness and clear inter-class separation. The resulting ranking is reported in Table~\ref{tab:gt}.

For each pre-trained model and each $n_{\mathrm{shot}}$ configuration, we compute three evaluation metrics: the proposed EGiME score, cross-validation classification accuracy with KNN (CV(KNN)), and cross-validation classification accuracy with SVM (CV(SVM)).

To investigate the stability of model evaluation approaches, we employ a quantitative metric known as ranking error ($RE$). Let $R=\{r_{ij}\}_q$ be the ranking matrix whose diagonal elements are $0$. Its element $r_{ij}=1$ when the $i$th model is superior to the $j$th one, and otherwise $r_{ij}=-1$. Let $R_g$ be the ground-truth ranking matrix and $R_e$ be the one obtained by an evaluation method. The ranking error is then defined as $RE=\|R_e-R_g\|_0/2$, where $\|\cdot\|_0$ denotes the 0-norm, i.e., the number of nonzero entries of the difference matrix divided by two.

\subsection{Agreement with the Ground-Truth Accuracy}

\begin{figure}[!htb]
\centering
\includegraphics[width=0.6\textwidth]{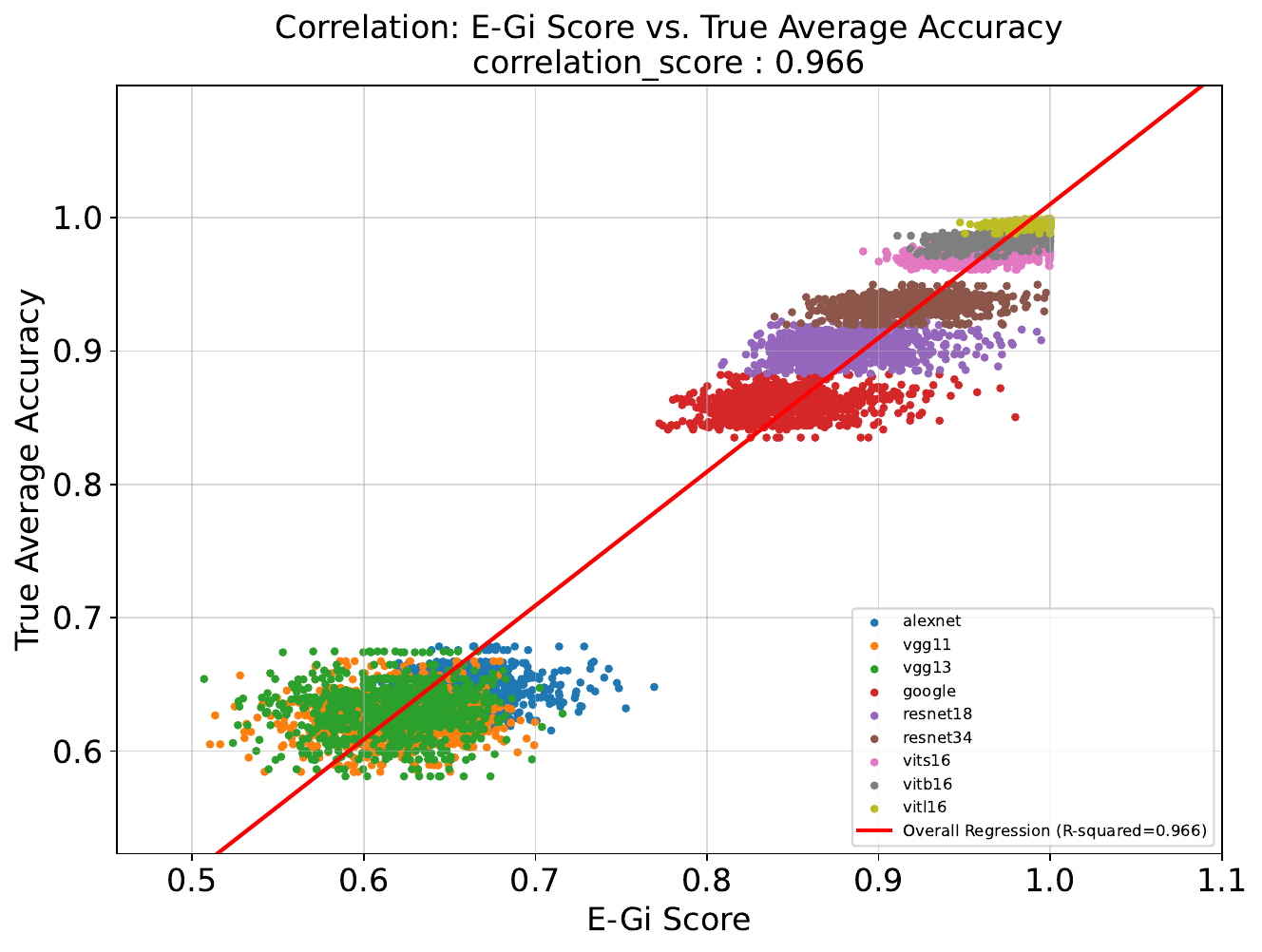}
\caption{EGiME as a low-label proxy for the three-predictor ground-truth accuracy. The overall regression score of $R^2=0.966$ indicates that EGiME preserves the accuracy trend and can serve as a practical pre-screening alternative to fully supervised model selection.}
\label{fig:corr}
\end{figure}

The practical role of Fig.~\ref{fig:corr} is to test whether EGiME can act as a low-label proxy for the fully supervised reference rather than merely to report a statistical coefficient. In a model-selection workflow, EGiME can first be computed with limited labels, weak backbones can be removed before expensive full-label evaluation, and only the remaining models need the three-predictor reference. The fitted trend in Fig.~\ref{fig:corr} reaches $R^2=0.966$, indicating strong agreement between the two evaluation procedures. Thus, on this benchmark, EGiME preserves both the accuracy trend and the relative ordering of models while requiring labels for only a small subset, making it a practical alternative to conventional supervised evaluation when full annotation is unavailable.

The plot also separates models according to representation quality: ViT-B/16 and ResNet34 consistently occupy the upper-right region associated with high EGiME scores and high accuracy, whereas AlexNet and VGG13 occupy the lower-left region. This separation is useful in practice because a high EGiME score is associated with stronger class compactness and clearer global separation, allowing weak backbones to be filtered out before full annotation. It also provides a geometric explanation for the ranking behavior examined below: the score is aligned with the fully labeled criterion while remaining computable from limited labels.

\subsection{Experimental Analysis of Model Ranking Stability in Few-Shot Scenarios}

To rigorously evaluate the model ranking stability and sample efficiency of the proposed weighted entropy method in few-shot scenarios, Table~\ref{tab:gt} first presents the ground-truth rankings of pre-trained models on five benchmark datasets, serving as the absolute reference anchor for all subsequent comparative experiments.

\begin{table}[htbp]
    \centering
    \caption{Ground-Truth Rankings of Pre-trained Models on Five Benchmark Datasets, Obtained by Averaging the Accuracies of K-means, KNN, and SVM}
    \label{tab:gt}
    \resizebox{\textwidth}{!}{%
    \begin{tabular}{lccccccccc}
        \toprule
        \textbf{Dataset} & \textbf{AlexNet} & \textbf{VGG11} & \textbf{VGG13} & \textbf{GoogLeNet} & \textbf{ResNet18} & \textbf{ResNet34} & \textbf{ViT-S} & \textbf{ViT-B} & \textbf{ViT-L} \\
        \midrule
        \textbf{Mini-val}   & 7 & 9 & 8 & 6 & 5 & 3 & 4 & 1 & 2 \\
        \textbf{STL-10}     & 7 & 8 & 9 & 6 & 5 & 4 & 3 & 2 & 1 \\
        \textbf{CIFAR-10}   & 7 & 8 & 9 & 4 & 5 & 3 & 6 & 1 & 2 \\
        \textbf{EuroSAT}    & 7 & 8 & 9 & 4 & 3 & 1 & 5 & 6 & 2 \\
        \textbf{CIFAR-100}  & 7 & 8 & 9 & 6 & 5 & 4 & 2 & 3 & 1 \\
        \bottomrule
    \end{tabular}
    }
\end{table}

In the specific evaluation process, for the five selected datasets (Mini-val, STL-10, CIFAR-10, EuroSAT, and CIFAR-100), we randomly sample 100 images per class to construct the base evaluation pool. To thoroughly eliminate random sampling bias and ensure the statistical significance of the results, for the proposed weighted entropy method, the two classical supervised baselines (CV(KNN) and CV(SVM)), and three recent transferability estimation metrics (LEEP, LogME, and PACTran), we independently repeat the model performance ranking experiments 100 times under different labeled sample proportions (i.e., extracting a small number of samples $n_{\text{shot}}$ per class). The results of all cycles are aggregated to obtain the corresponding evaluation score of each model under each labeling proportion. It is worth noting that in each independent sampling, apart from the extremely small number of $n_{\text{shot}}$ labeled samples used for evaluation guidance or baseline classifier training, the remaining massive amount of unlabeled samples in the experimental pool (accounting for 80\% $\sim$ 96\%) are completely retained and utilized to construct the global clustering manifold topological basis, thereby providing solid unsupervised statistical support for the calculation of weighted entropy.

For the three transferability metrics, each is adopted in its original formulation: LEEP~\cite{nguyen2020leep} is computed with a real ImageNet-1000 source classifier $\theta$, LogME~\cite{you2021logme} with the evidence of a Bayesian linear probe maximized over $(\alpha,\beta)$ by a fixed-point iteration, and PACTran~\cite{ding2022pactran} with a PAC-Bayes bound instantiated under a Gaussian prior. Where the definition of a metric requires a labeled target set, that set is taken to be the $K$-shot labeled subset.

It should be stressed that these three metrics are recent \emph{general-purpose} transferability estimators: they were proposed and validated on a fully labeled target set, none of them was designed for the few-shot regime, and their behavior when only a handful of labeled samples per class is available has not been examined before. The experiment reported below is therefore, to our knowledge, the first to place them under a $K$-shot budget.

In the quantitative phase of the model ranking task, we adopt ``Average Ranking Mistakes'' as the core evaluation metric. The specific operation is as follows: in each independent run, the rankings derived from the weighted entropy evaluation, the predicted accuracy-based rankings of CV(KNN) and CV(SVM), and the score-based rankings of LEEP, LogME, and PACTran are individually compared with the ground-truth rankings in Table~\ref{tab:gt}. We precisely count the number of model pairs that exhibit inversions (inconsistencies) between the evaluated ranking and the ground-truth ranking, and finally average the number of mistakes over the 100 independent experiments. Table~\ref{tab:rm} systematically presents the average ranking mistakes of all six methods on the five datasets under different labeling proportions.

\begin{table}[!htb]
    \centering
    \caption{Average Ranking Mistakes on Five Datasets Under Different Labeling Proportions}
    \label{tab:rm}
    \renewcommand{\arraystretch}{0.85}
    \resizebox{\textwidth}{!}{%
    \begin{tabular}{llccccccccc}
        \toprule
        \multirow{2}{*}{\textbf{Dataset}} & \multirow{2}{*}{\textbf{Method}} & \multicolumn{9}{c}{\textbf{Labels Per Class ($n_{\mathrm{shot}}$)}} \\
        \cmidrule{3-11}
        & & \textbf{4} & \textbf{6} & \textbf{8} & \textbf{10} & \textbf{12} & \textbf{14} & \textbf{16} & \textbf{18} & \textbf{20} \\
        \midrule
        \multirow{6}{*}{\textbf{Mini-val}}
        & CV(KNN) & 3.01 & 2.88 & 2.72 & 2.43 & 2.25 & 2.33 & 2.23 & 1.92 & 1.81 \\
        & CV(SVM) & 3.38 & 3.74 & 4.56 & 4.70 & 4.82 & 4.59 & 4.70 & 4.64 & 4.53 \\
        & LEEP & 12.59 & 11.91 & 12.04 & 12.22 & 11.82 & 11.80 & 11.83 & 11.54 & 11.28 \\
        & LogME & 3.87 & 3.96 & 4.14 & 4.12 & 4.29 & 4.26 & 4.30 & 4.22 & 4.24 \\
        & PACTran & 8.23 & 8.78 & 9.24 & 9.31 & 9.42 & 9.52 & 9.62 & 9.71 & 9.74 \\
        & EGiME & \underline{2.87} & \underline{2.62} & \underline{2.09} & \underline{2.00} & \underline{1.92} & \underline{1.83} & \underline{1.70} & \underline{1.54} & \underline{1.38} \\
        \midrule
        \multirow{6}{*}{\textbf{STL-10}}
        & CV(KNN) & 2.84 & 2.24 & 2.01 & 1.86 & 1.49 & 1.39 & 1.41 & 1.27 & 1.25 \\
        & CV(SVM) & 3.14 & 2.70 & 2.68 & 2.92 & 2.57 & 2.71 & 2.61 & 2.59 & 2.69 \\
        & LEEP & 9.56 & 8.75 & 8.41 & 8.06 & 7.74 & 7.60 & 7.51 & 7.37 & 7.36 \\
        & LogME & \underline{2.09} & \underline{1.64} & 1.79 & 1.85 & 1.87 & 2.25 & 2.26 & 2.45 & 2.51 \\
        & PACTran & 8.86 & 9.68 & 9.92 & 9.99 & 9.97 & 10.07 & 10.41 & 10.75 & 11.12 \\
        & EGiME & 2.14 & 1.71 & \underline{1.60} & \underline{1.37} & \underline{1.30} & \underline{1.29} & \underline{1.28} & \underline{1.11} & \underline{1.18} \\
        \midrule
        \multirow{6}{*}{\textbf{CIFAR-10}}
        & CV(KNN) & 5.67 & 4.71 & 4.14 & 3.93 & 3.67 & 3.17 & 3.54 & 2.88 & 2.82 \\
        & CV(SVM) & 5.54 & 4.77 & 4.37 & 4.13 & 3.83 & 3.47 & 4.17 & 3.75 & 3.70 \\
        & LEEP & 11.34 & 10.95 & 10.91 & 11.16 & 10.91 & 10.92 & 10.91 & 10.88 & 10.81 \\
        & LogME & 6.02 & 4.99 & 4.67 & 4.30 & 4.29 & 4.17 & 4.00 & 3.78 & 3.75 \\
        & PACTran & 12.21 & 12.14 & 12.40 & 12.45 & 12.32 & 12.55 & 12.88 & 12.92 & 13.04 \\
        & EGiME & \underline{3.66} & \underline{3.48} & \underline{3.39} & \underline{3.11} & \underline{2.83} & \underline{2.28} & \underline{2.75} & \underline{2.40} & \underline{2.15} \\
        \midrule
        \multirow{6}{*}{\textbf{EuroSAT}}
        & CV(KNN) & 7.49 & 7.54 & 6.42 & 6.57 & 6.94 & 6.76 & 6.30 & 6.02 & 6.06 \\
        & CV(SVM) & 8.16 & 8.12 & 8.35 & 8.78 & 9.13 & 9.50 & 10.16 & 9.83 & 10.70 \\
        & LEEP & 17.15 & 16.17 & 15.46 & 15.24 & 15.19 & 14.99 & 14.98 & 14.72 & 14.44 \\
        & LogME & 14.49 & 9.41 & 7.08 & 6.28 & 6.31 & 6.02 & 6.81 & 7.15 & 6.69 \\
        & PACTran & 7.47 & 7.79 & 8.10 & 8.27 & 8.33 & 8.33 & 8.36 & 8.55 & 8.50 \\
        & EGiME & \underline{5.86} & \underline{5.13} & \underline{5.27} & \underline{5.11} & \underline{4.53} & \underline{4.69} & \underline{4.90} & \underline{4.66} & \underline{4.45} \\
        \midrule
        \multirow{6}{*}{\textbf{CIFAR-100}}
        & CV(KNN) & 1.66 & 1.14 & 1.22 & 1.16 & 1.11 & 1.19 & 1.12 & 1.09 & 1.21 \\
        & CV(SVM) & 1.60 & 1.46 & 1.58 & 2.03 & 2.64 & 3.26 & 3.32 & 4.05 & 3.91 \\
        & LEEP & 11.73 & 14.03 & 14.58 & 14.93 & 15.48 & 16.32 & 16.99 & 17.00 & 17.00 \\
        & LogME & 12.60 & 11.96 & 10.52 & 10.32 & 10.22 & 10.02 & 9.98 & 9.77 & 9.74 \\
        & PACTran & 8.23 & 8.33 & 8.52 & 8.99 & 9.22 & 9.51 & 9.98 & 9.85 & 10.23 \\
        & EGiME & \underline{0.95} & \underline{0.83} & \underline{0.62} & \underline{0.51} & \underline{0.52} & \underline{0.57} & \underline{0.37} & \underline{0.33} & \underline{0.34} \\
        \bottomrule
    \end{tabular}
    }
\end{table}

\begin{figure}[!htb]
\centering
\begin{minipage}[b]{0.31\linewidth}
\centering
\includegraphics[width=\linewidth]{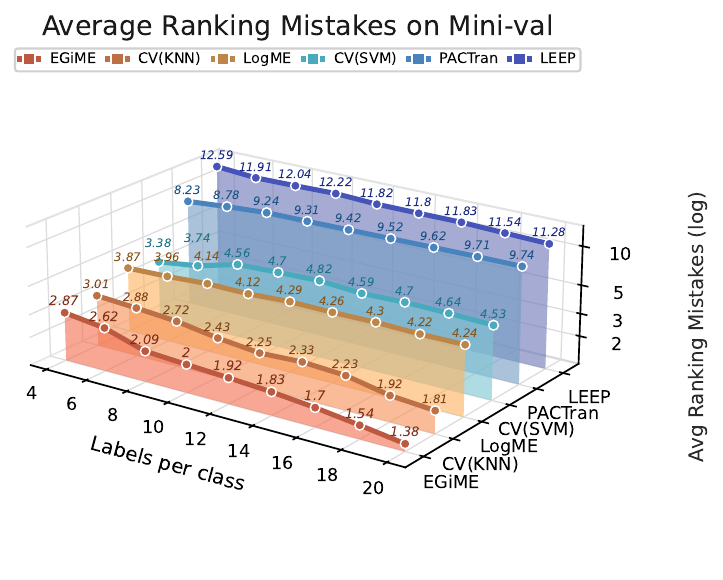}
\centerline{\small (A) Mini-val}
\end{minipage}\hfill
\begin{minipage}[b]{0.31\linewidth}
\centering
\includegraphics[width=\linewidth]{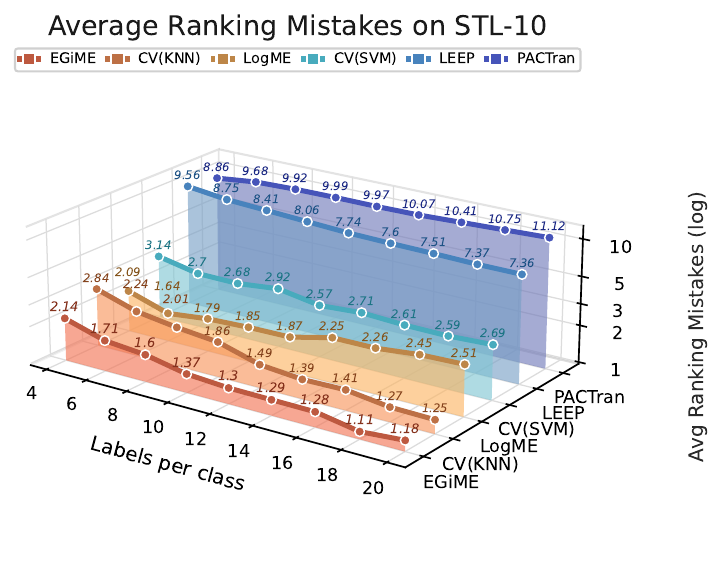}
\centerline{\small (B) STL-10}
\end{minipage}

\vspace{4pt}
\begin{minipage}[b]{0.31\linewidth}
\centering
\includegraphics[width=\linewidth]{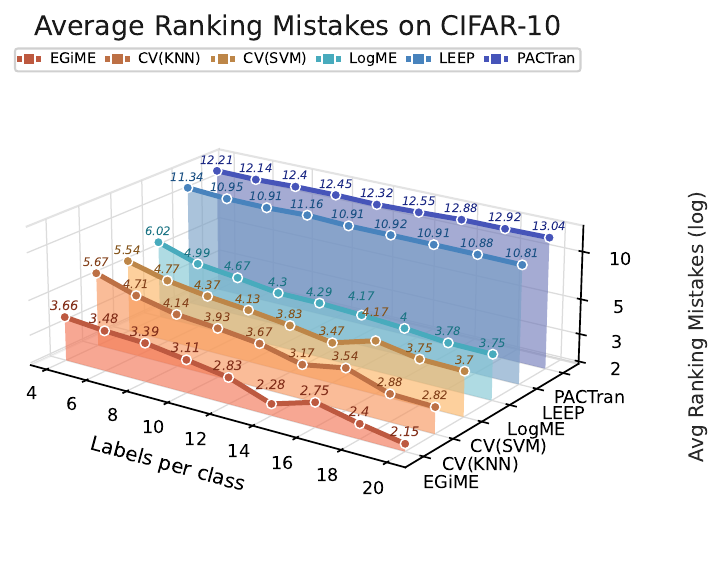}
\centerline{\small (C) CIFAR-10}
\end{minipage}\hfill
\begin{minipage}[b]{0.31\linewidth}
\centering
\includegraphics[width=\linewidth]{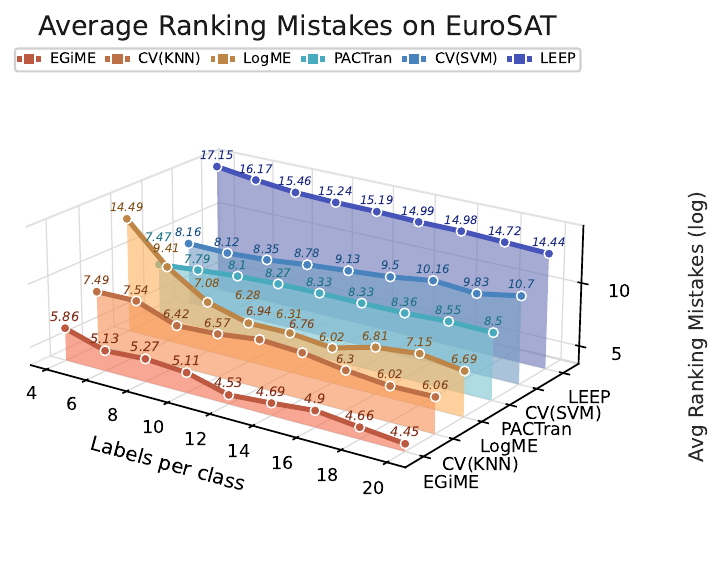}
\centerline{\small (D) EuroSAT}
\end{minipage}\hfill
\begin{minipage}[b]{0.31\linewidth}
\centering
\includegraphics[width=\linewidth]{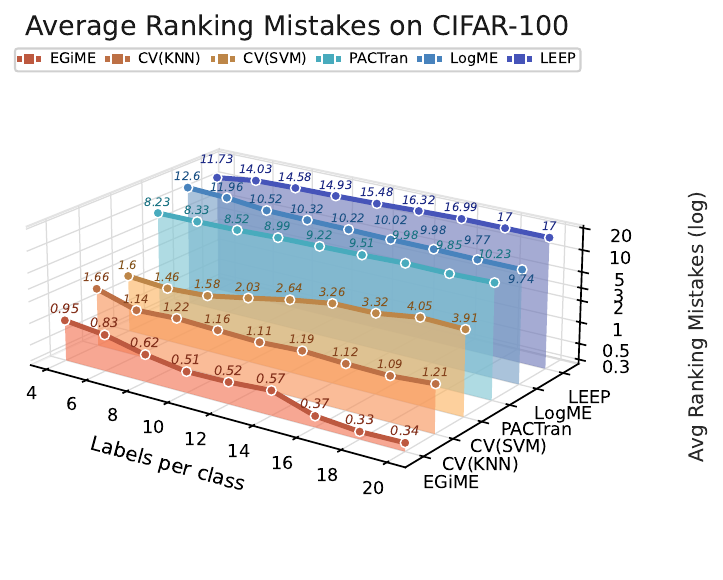}
\centerline{\small (E) CIFAR-100}
\end{minipage}
\caption{Average ranking mistakes of the six methods as a function of the number of labeled samples per class, drawn as waterfall surfaces; the vertical axis is logarithmic, since LEEP and PACTran exceed the remaining methods by up to an order of magnitude. Within each panel the methods run from front (EGiME) to back (LEEP).}
\label{fig:waterfall}
\end{figure}

Table~\ref{tab:rm} shows that, compared with CV(KNN) and CV(SVM), EGiME yields lower average ranking mistakes on all five datasets and at every labeling proportion considered. This indicates that, in the tested few-shot settings, using a small number of labeled samples as anchors while exploiting the unlabeled pool through clustering provides more stable model rankings than the two cross-validation baselines. Fig.~\ref{fig:waterfall} further visualizes the average ranking mistakes of all six methods on the five datasets.

Beyond the two supervised baselines, three recent transferability estimation metrics, LEEP~\cite{nguyen2020leep}, LogME~\cite{you2021logme}, and PACTran~\cite{ding2022pactran}, are included as supplementary comparisons. These metrics assign each model a single transferability score and rank the models by that score. As shown in Table~\ref{tab:rm}, LEEP and PACTran generally produce higher average ranking mistakes, whereas LogME is more competitive but does not consistently outperform EGiME; in particular, LogME is lower than EGiME on STL-10 at 4 and 6 shots. These results should not be interpreted as evidence that the three metrics are generally ineffective, because they were originally designed and evaluated with fully labeled target sets. Their comparison here is therefore limited to ranking performance under the few-shot budget.

Across the 100 repetitions, CV(KNN) and CV(SVM) show larger run-to-run fluctuations, whereas EGiME generally has a lower and more concentrated error distribution. These results suggest that EGiME is less sensitive to random few-shot sampling than the two cross-validation baselines.

Across the tested labeling proportions, EGiME remains below both cross-validation baselines in most settings. The gap is generally larger in the low-label regime, while the curves approach one another on some datasets as the number of labels increases. This trend is not monotonic on every dataset, so it is interpreted as a descriptive pattern rather than evidence of asymptotic convergence.

To examine sample efficiency, we analyze CIFAR-100, which has high feature dimensionality and fine-grained categories. From 4 to 20 shots, the average ranking mistake of EGiME decreases overall from 0.95 to 0.34, with small local increases at intermediate budgets. CV(KNN) changes only modestly from 1.66 to 1.21, while CV(SVM) increases from 1.60 to 3.91. These results suggest that EGiME benefits from additional labeled anchors, although the improvement is not strictly monotonic.

These results are consistent with the known difficulty of purely supervised classifiers in few-shot regimes~\cite{perez2021truefewshot,wang2020fslsurvey}. With very few training samples, SVM decision boundaries can be sensitive to sampling noise~\cite{cervantes2020svm}. EGiME instead uses the unlabeled pool to regularize the score through the global cluster structure. On CIFAR-10, STL-10, and Mini-val, the errors of most methods tend to decrease as more labels are used, while EGiME remains lower than the two cross-validation baselines in the tested settings.

To test generalization under domain shift, we evaluate EuroSAT. Because the pre-trained models originate from natural images, their feature spaces are less aligned with remote-sensing images, and ranking mistakes are generally higher than on the natural-image datasets. The EGiME error ranges from 4.45 to 5.86, compared with 6.06 to 7.54 for CV(KNN) and 8.12 to 10.70 for CV(SVM).

Across all nine labeling proportions on EuroSAT, EGiME has the lowest average ranking mistake among the six methods. This result supports its usefulness under the evaluated domain shift, although the experiment covers one target domain and does not by itself establish general cross-domain robustness.

\begin{figure}[!htb]
\centering
\includegraphics[width=0.30\textwidth]{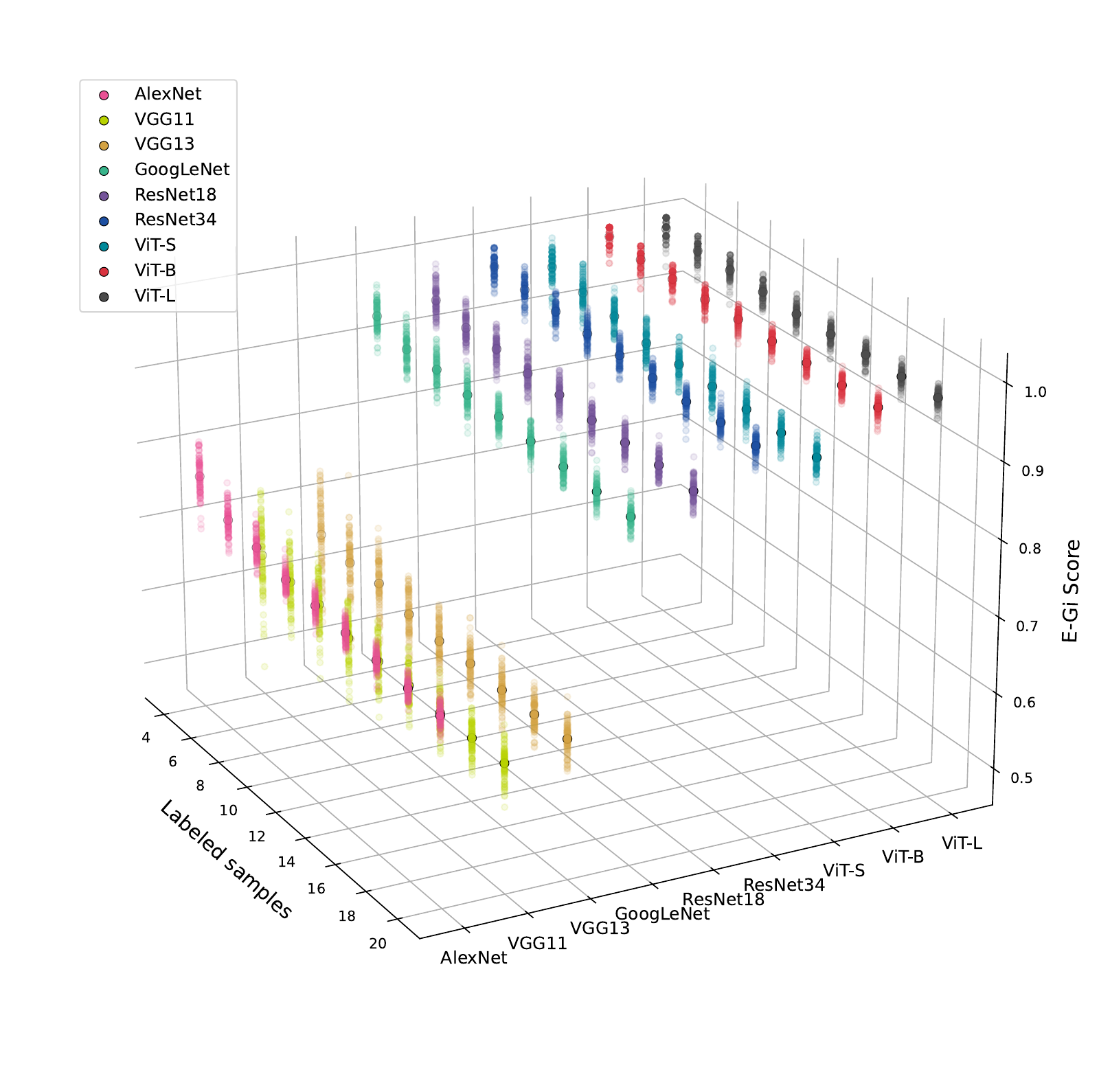}\hfill
\includegraphics[width=0.30\textwidth]{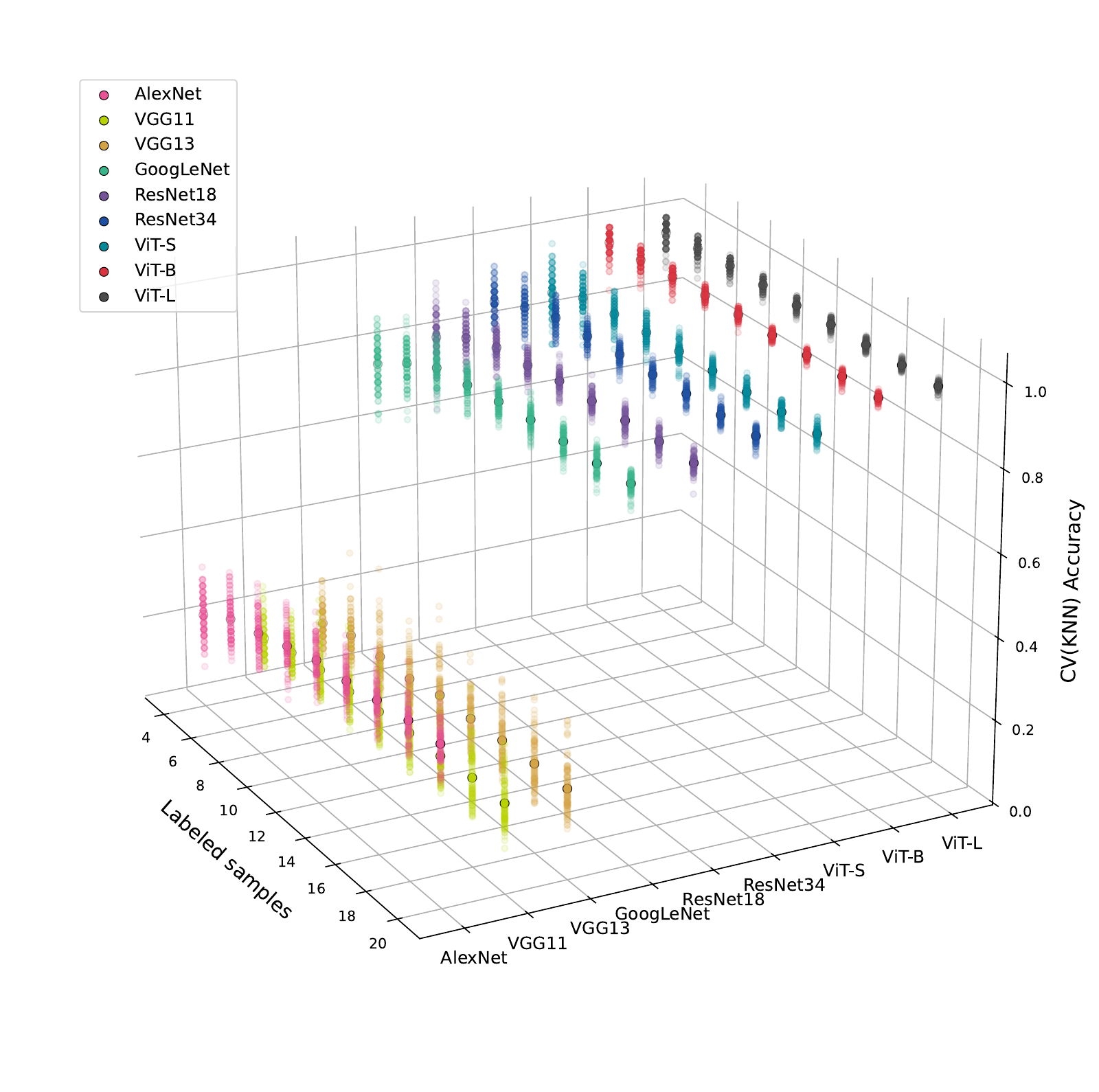}\hfill
\includegraphics[width=0.30\textwidth]{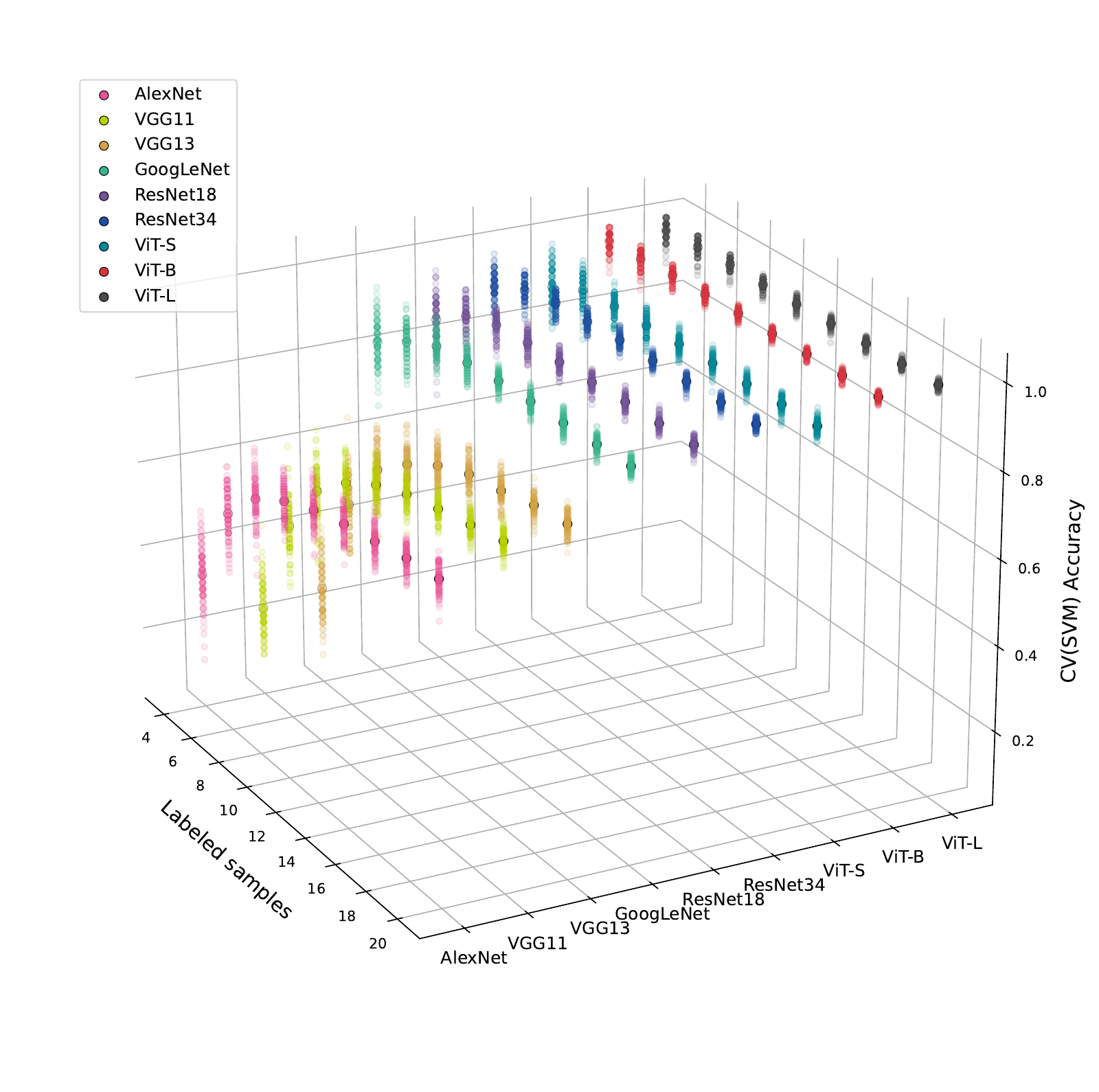}
\\[-0.1cm]
{\small (A) Mini-ImageNet}
\\[0.15cm]
\includegraphics[width=0.30\textwidth]{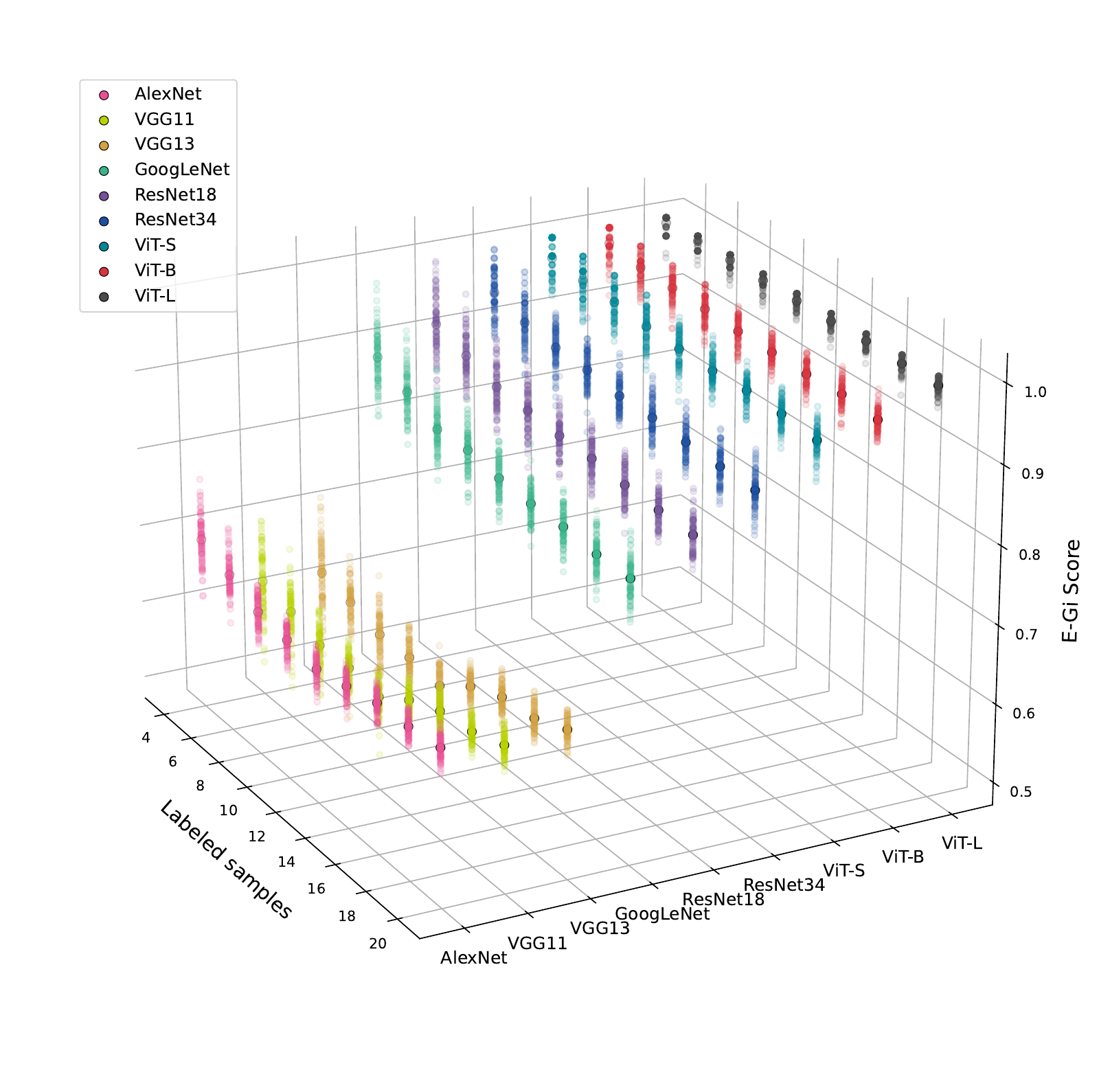}\hfill
\includegraphics[width=0.30\textwidth]{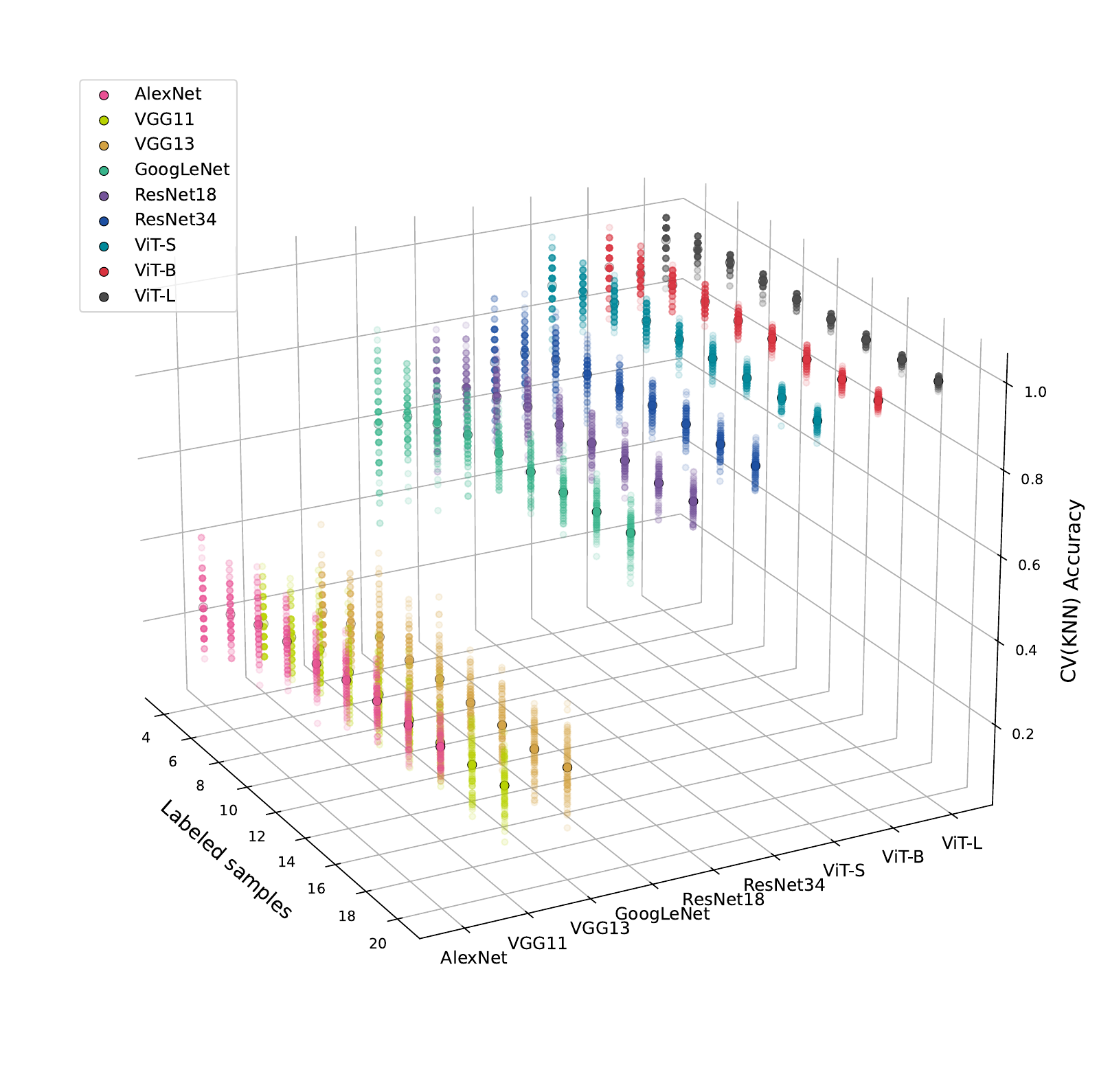}\hfill
\includegraphics[width=0.30\textwidth]{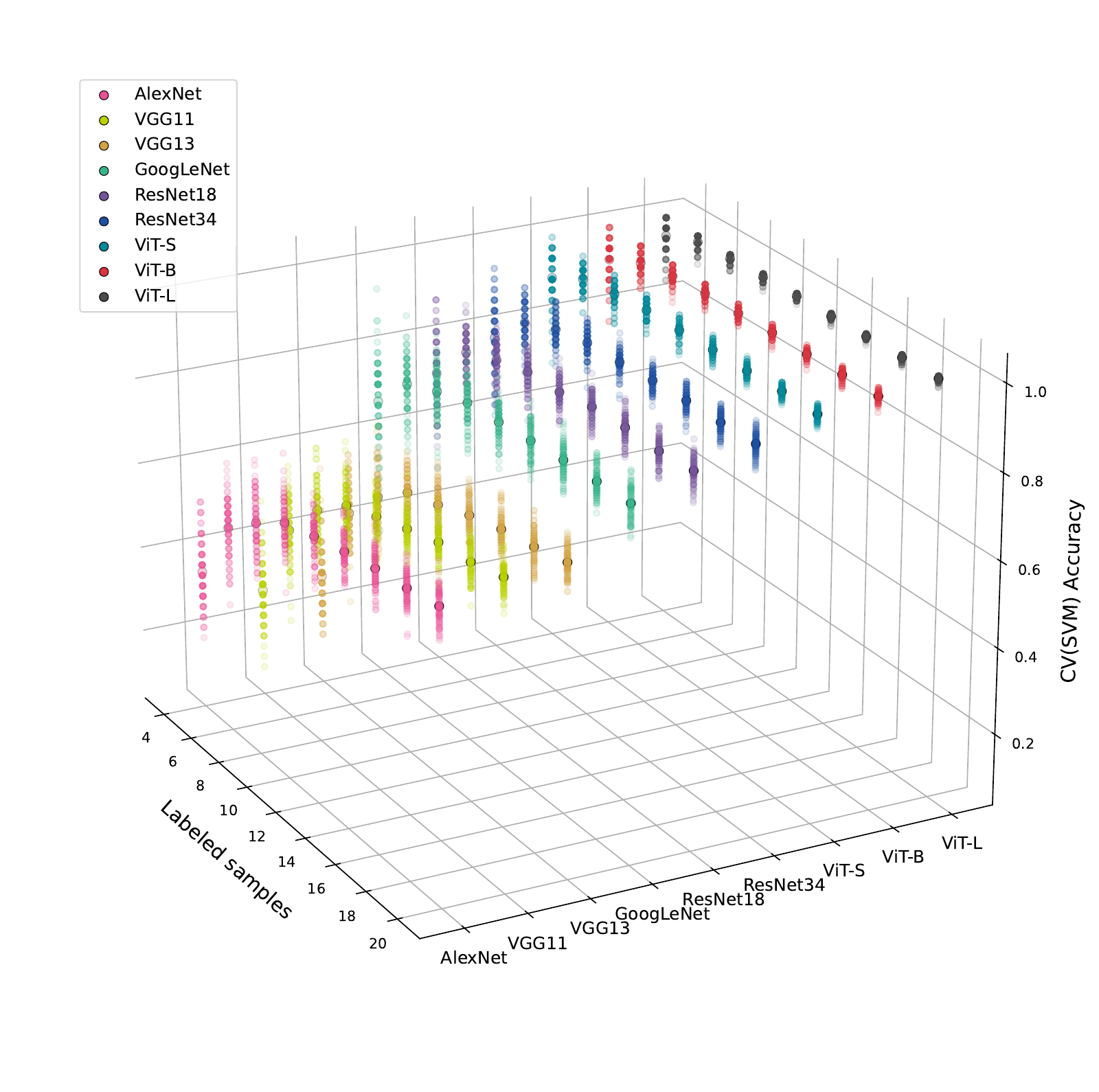}
\\[-0.1cm]
{\small (B) STL-10}
\\[0.15cm]
\includegraphics[width=0.30\textwidth]{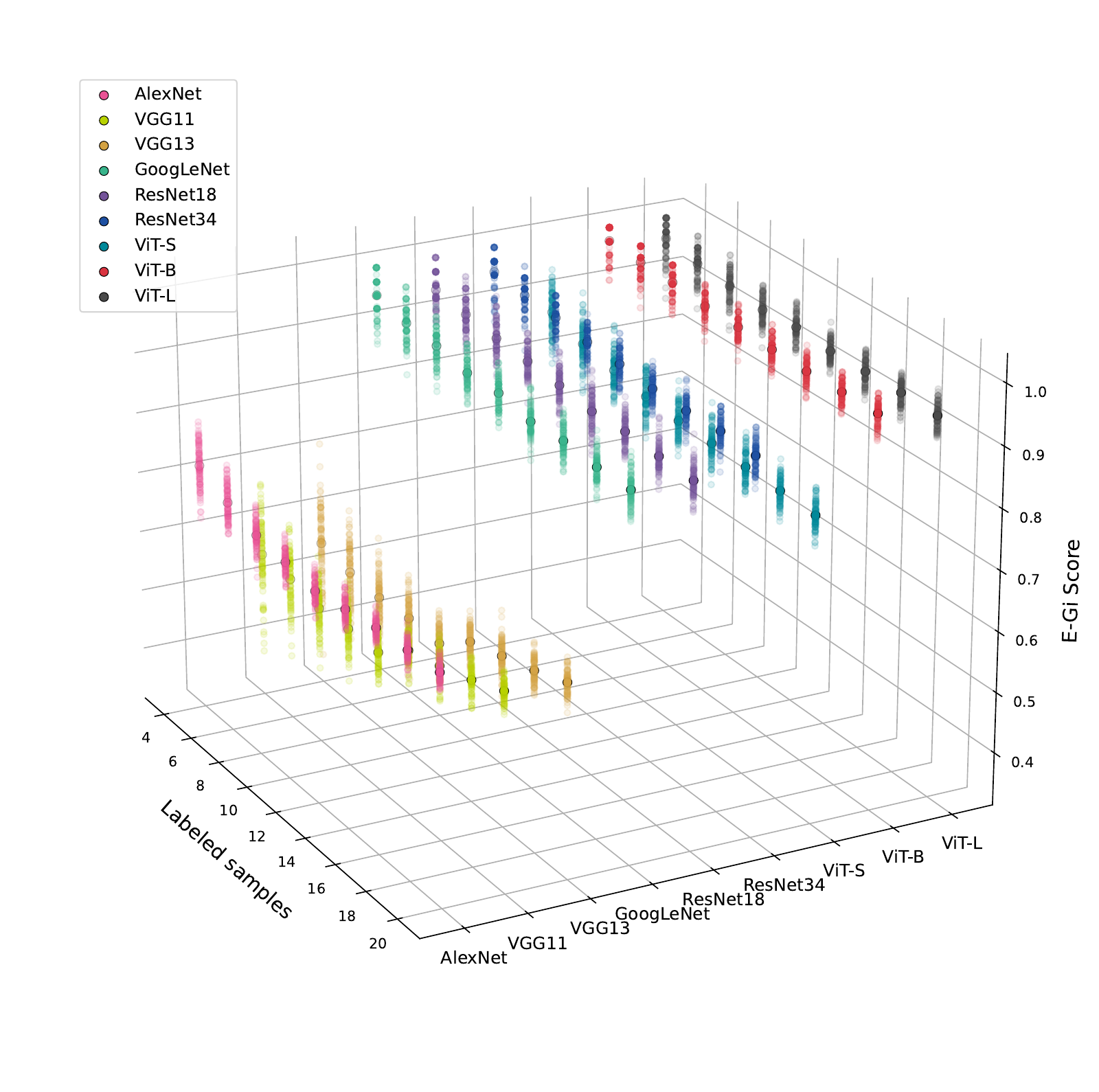}\hfill
\includegraphics[width=0.30\textwidth]{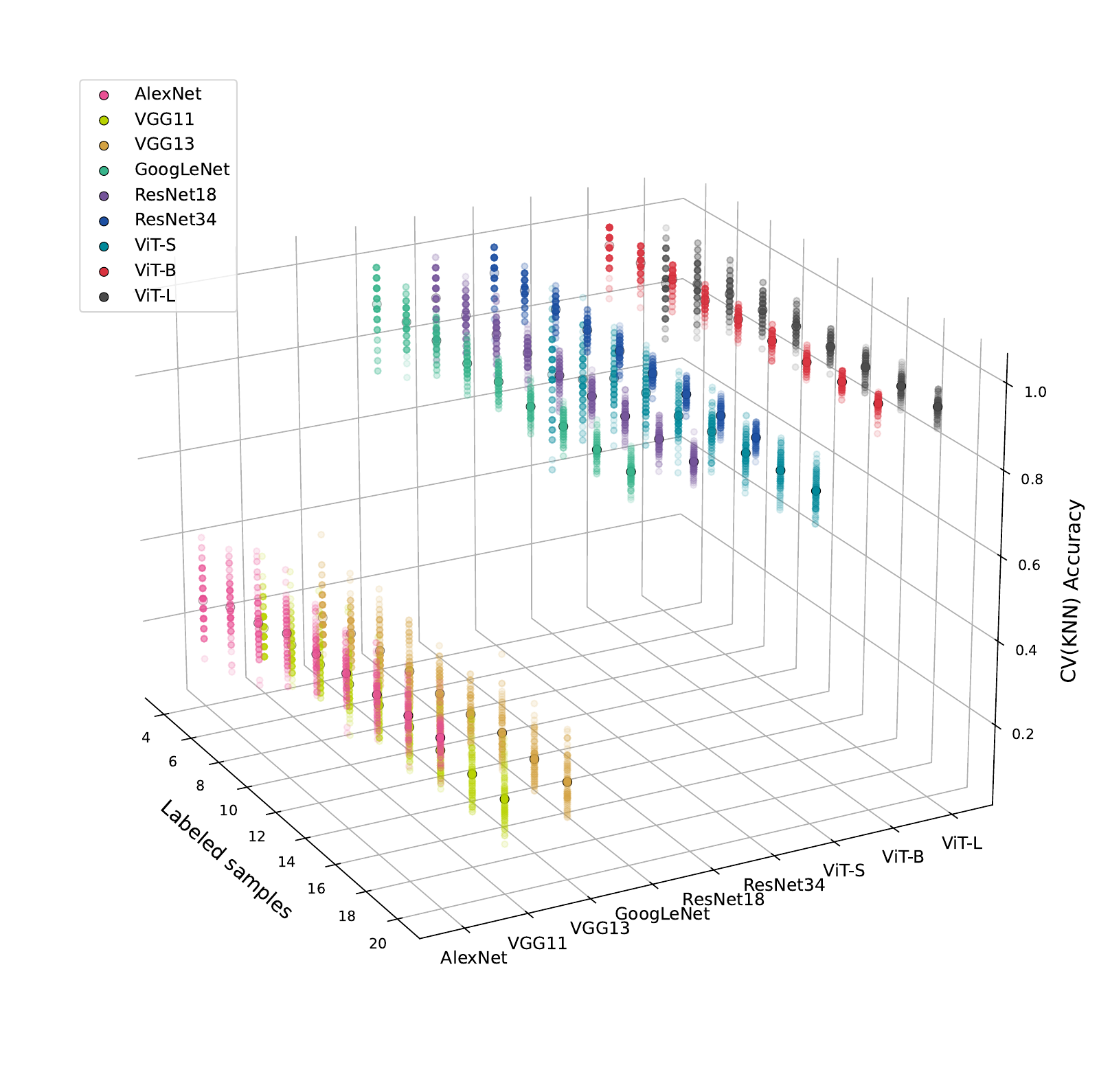}\hfill
\includegraphics[width=0.30\textwidth]{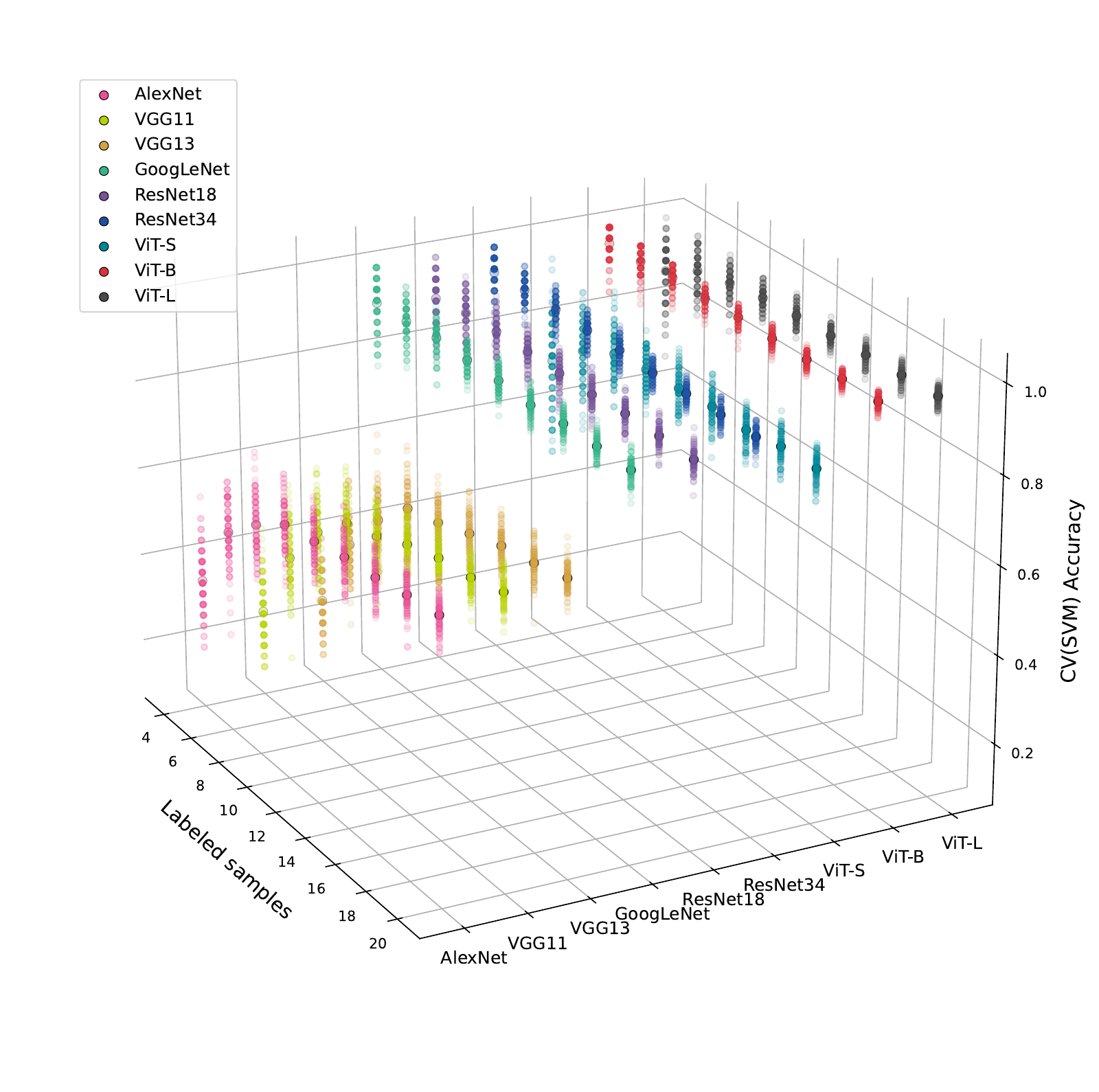}
\\[-0.1cm]
{\small (C) CIFAR-10}
\caption{Score distributions across different labeled sample sizes. (A) Mini-ImageNet, (B) STL-10, (C) CIFAR-10. Left: EGiME, Middle: CV(KNN), Right: CV(SVM).}
\label{fig:dist1}
\end{figure}

\begin{figure}[!htb]
\centering
\includegraphics[width=0.30\textwidth]{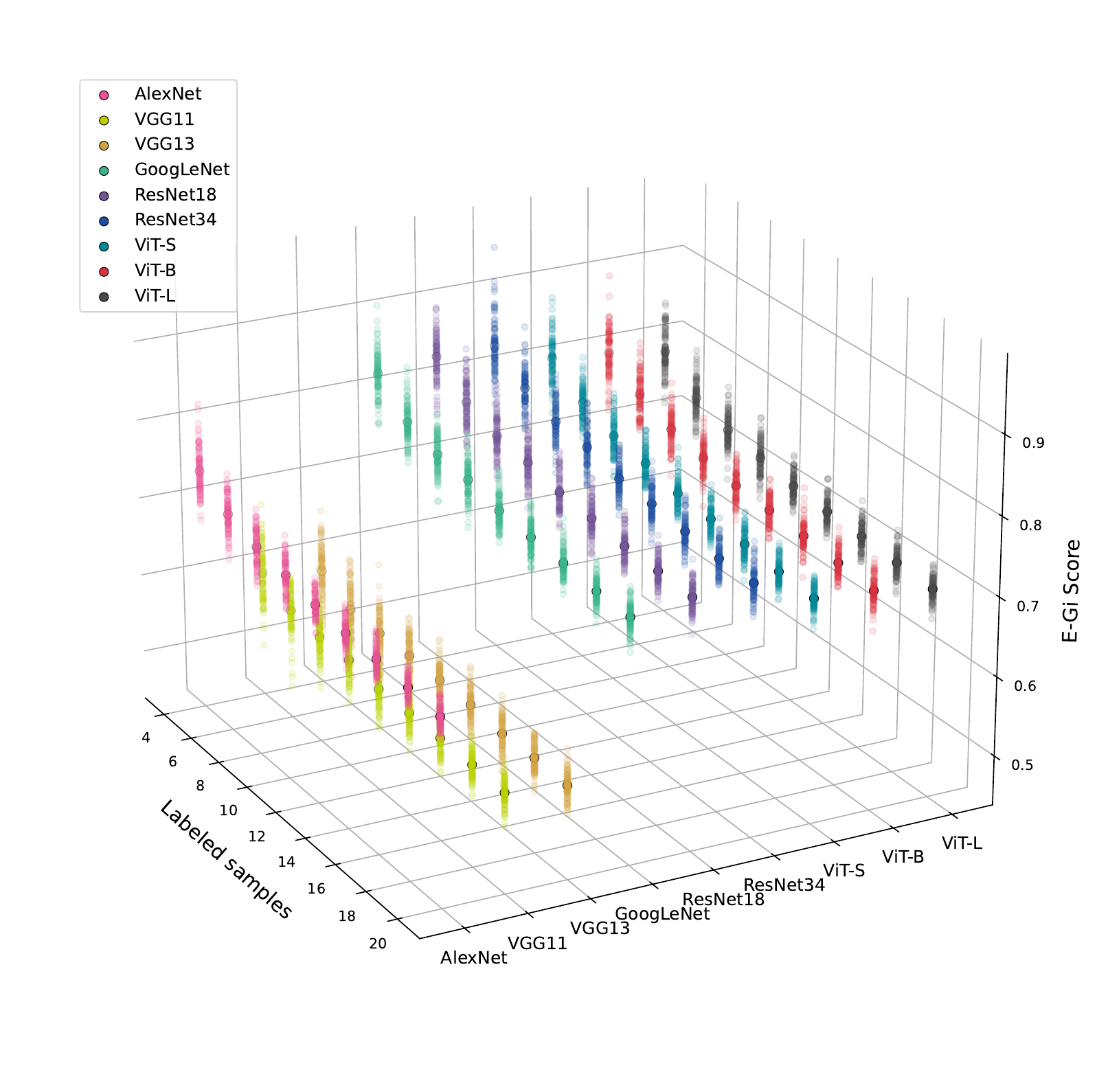}\hfill
\includegraphics[width=0.30\textwidth]{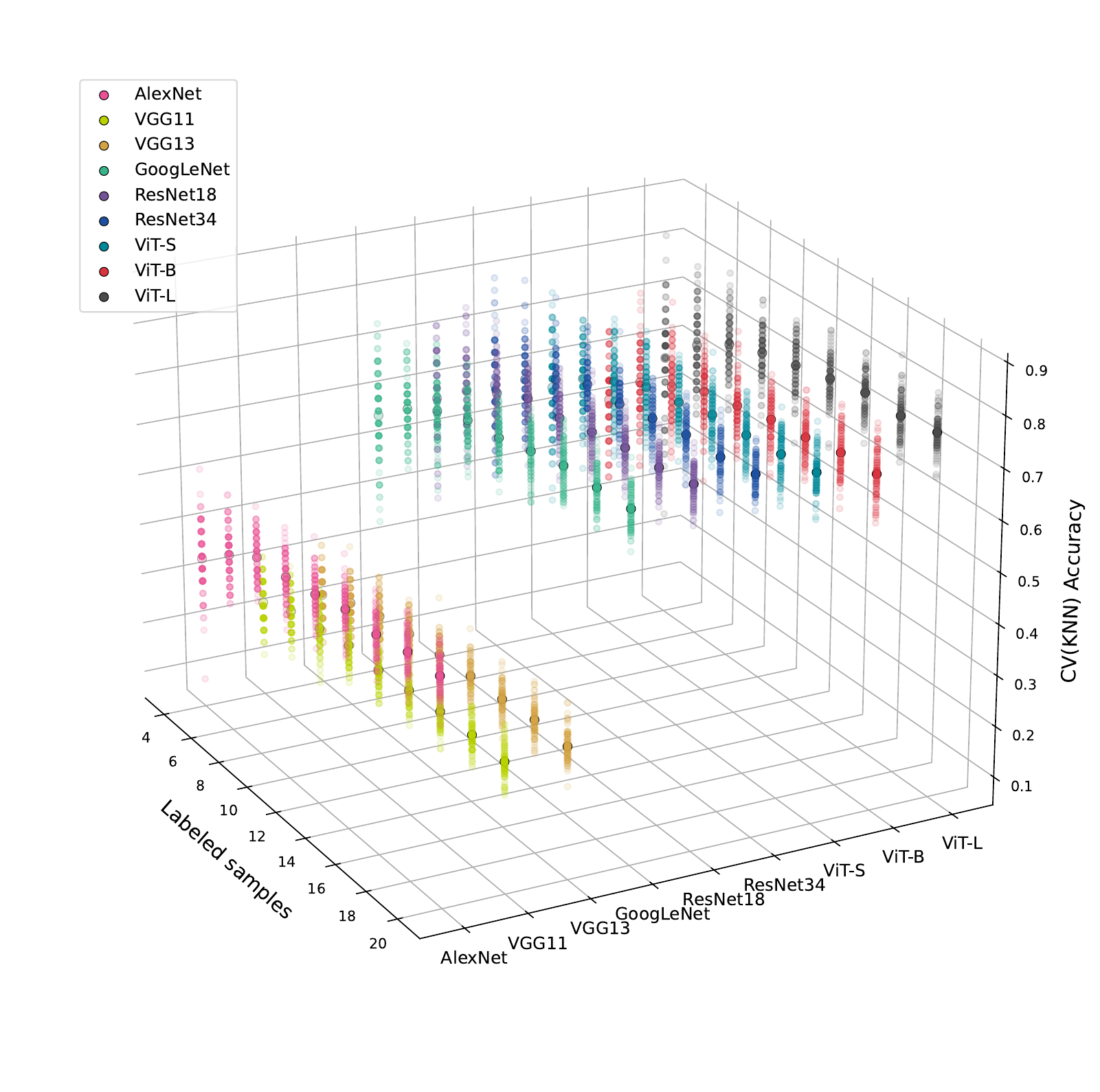}\hfill
\includegraphics[width=0.30\textwidth]{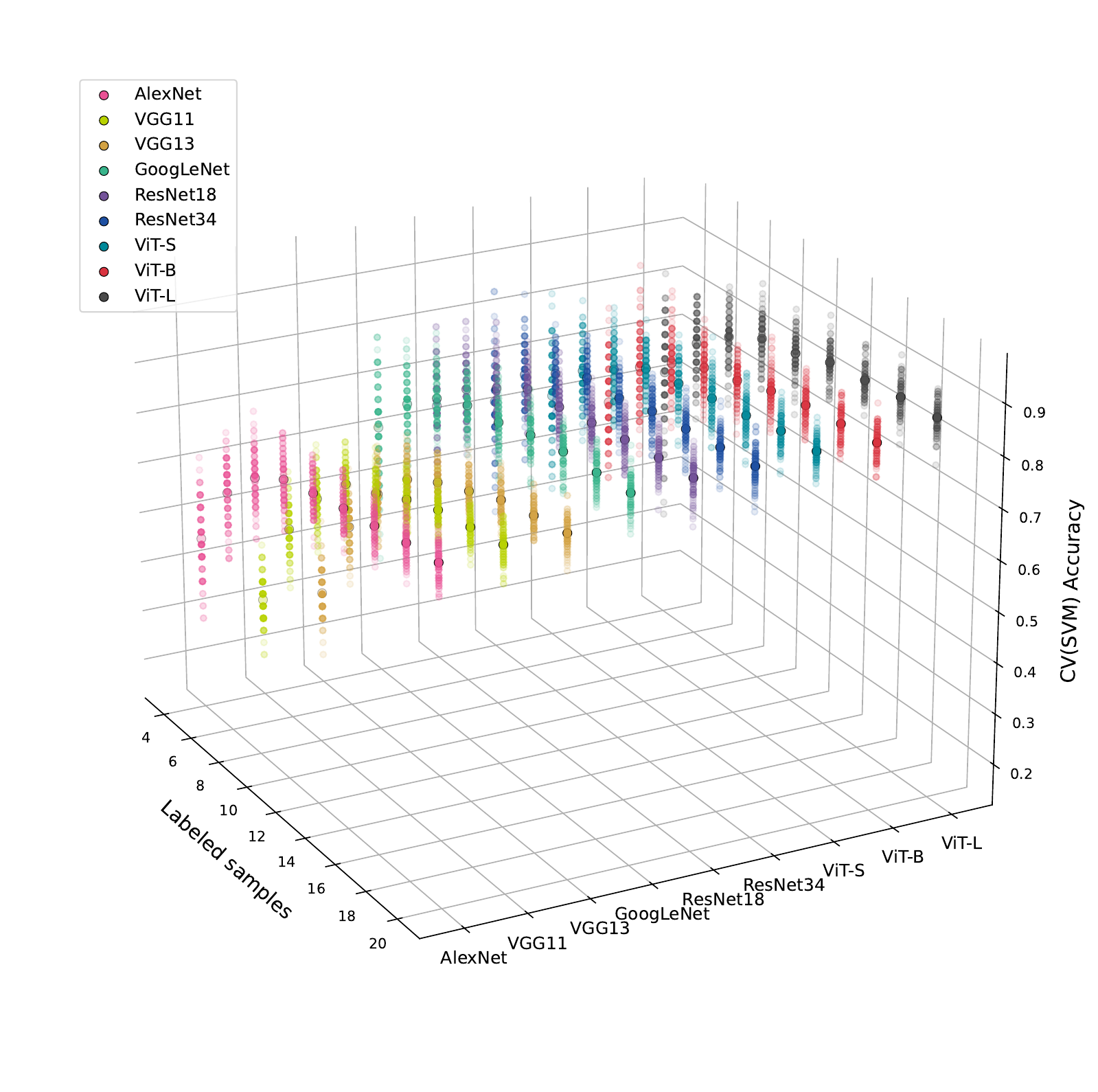}
\\[-0.1cm]
{\small (D) EuroSAT}
\\[0.15cm]
\includegraphics[width=0.30\textwidth]{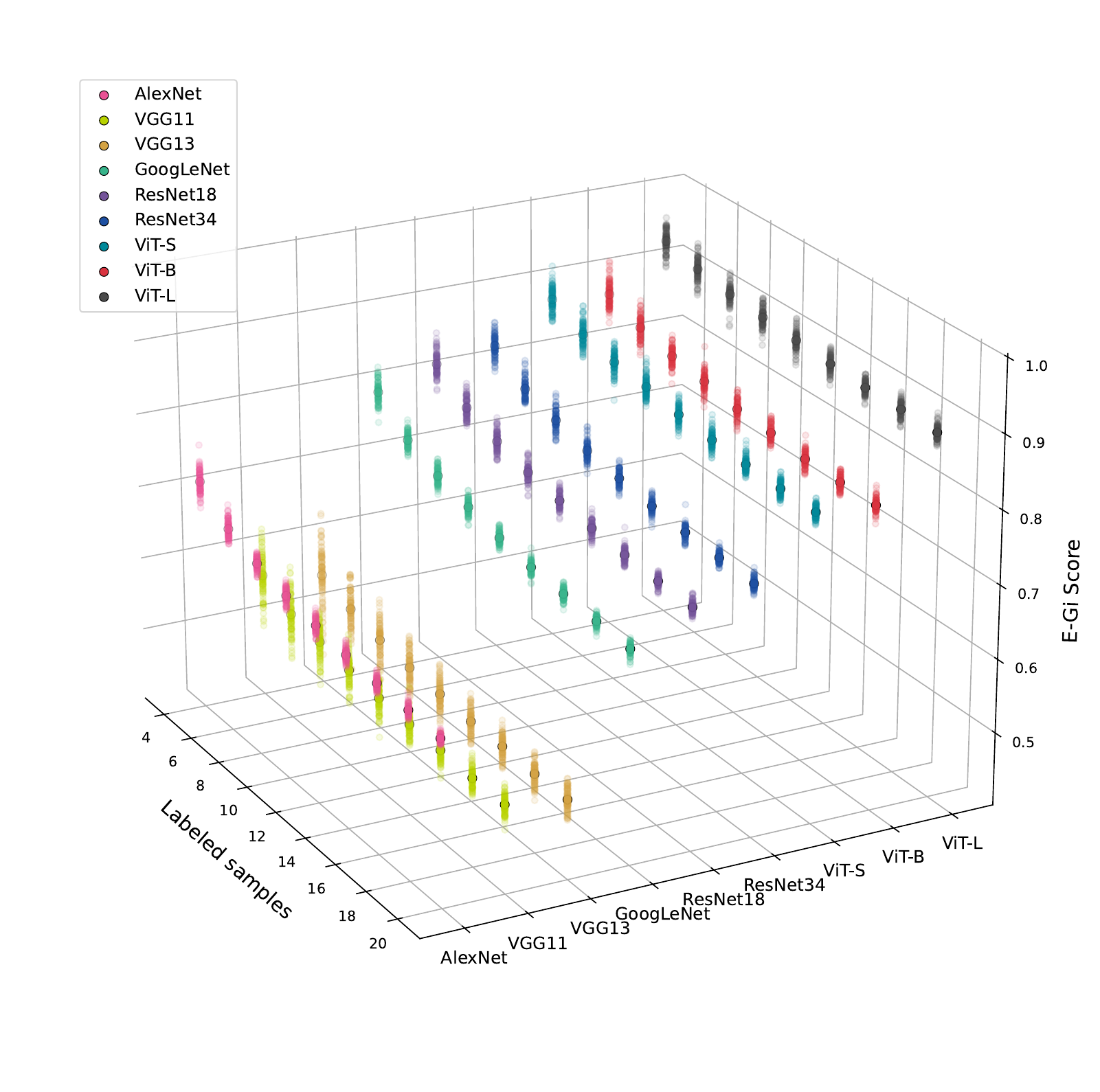}\hfill
\includegraphics[width=0.30\textwidth]{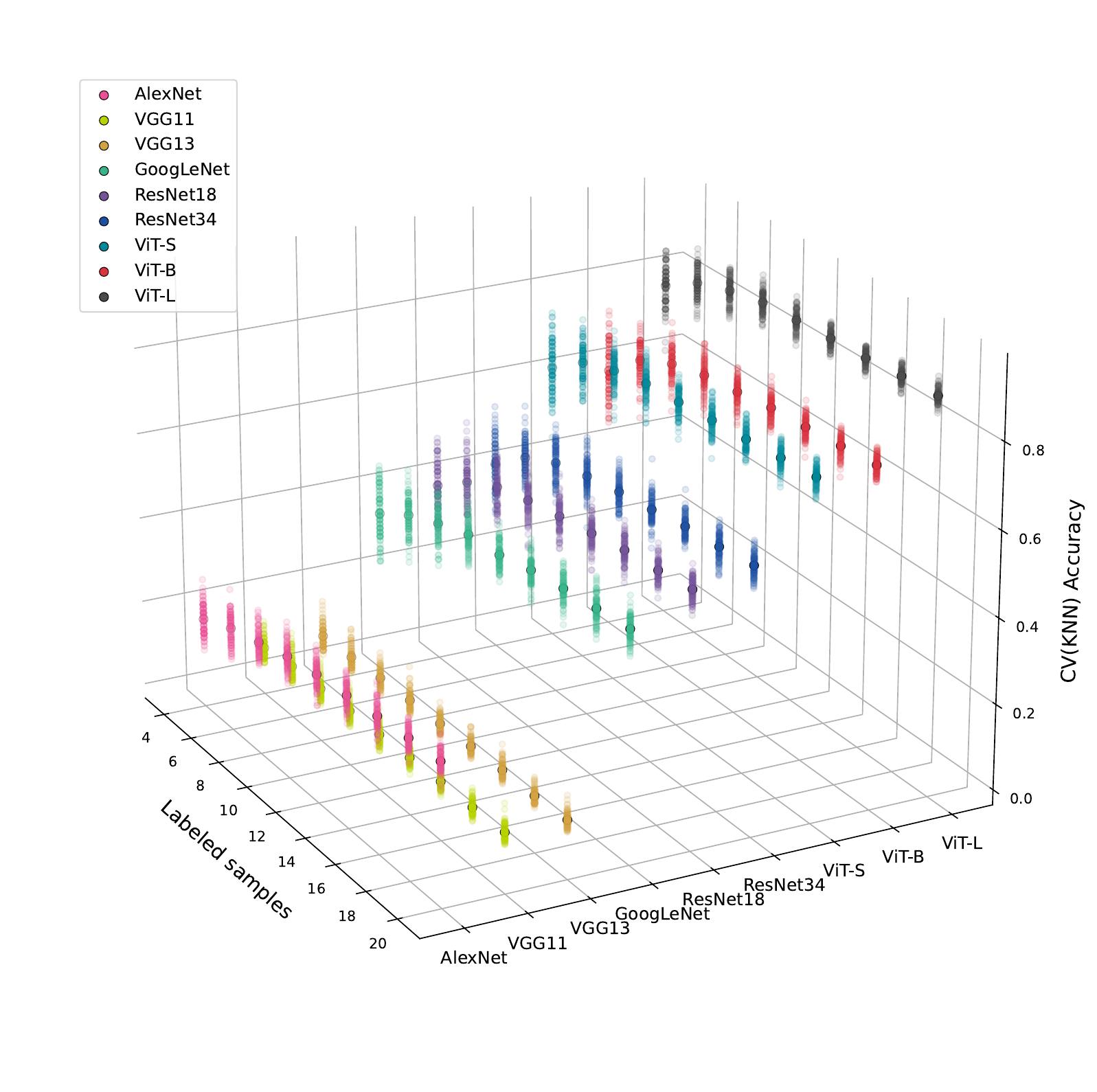}\hfill
\includegraphics[width=0.30\textwidth]{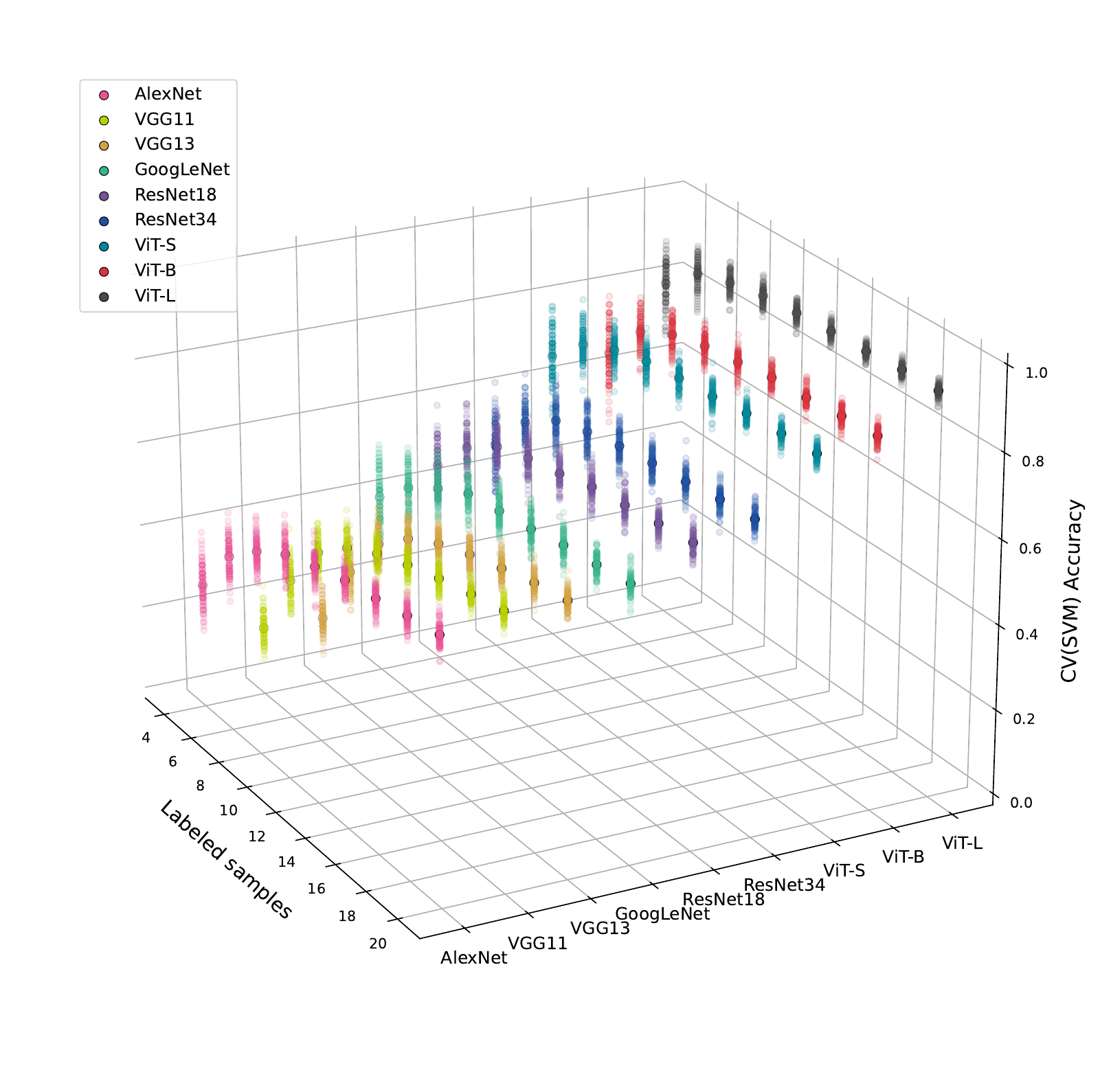}
\\[-0.1cm]
{\small (E) CIFAR-100}
\caption{Score distributions across different labeled sample sizes (continued). (D) EuroSAT, (E) CIFAR-100.}
\label{fig:dist2}
\end{figure}

The ranking comparison shows whether a method orders the candidate models correctly; it does not show whether the score assigned to an individual model is stable across labeling conditions. Fig.~\ref{fig:dist1} and Fig.~\ref{fig:dist2} present the per-model score distributions of EGiME, CV(KNN), and CV(SVM) on the five datasets. The three transferability metrics are not included because their scores are defined on different scales. The distributions show that EGiME scores are generally more concentrated across labeling proportions than those of the two baselines in the examined settings. These figures are therefore used as descriptive evidence of score stability.

\subsection{Ablation Experiment}

To verify the critical role of each component in the EGiME framework, we design two ablation experiments: (A) replacing the K-means clustering with equal partition to assess the contribution of the structural prior, and (B) decomposing the weighted entropy formula to assess the respective contributions of the inter-cluster and intra-cluster entropy terms.

\textbf{A. Clustering-based partitioning strategy.} We replace the K-means clustering module in the EGiME framework with an ``Equal Partition'' strategy as the baseline. The equal partition approach disregards the geometric structure and semantic distribution of the feature space, instead uniformly dividing samples into equally sized groups based on a fixed rule. Since the subsequent entropy-based evaluation formula remains identical under both strategies, this ablation experiment effectively isolates and quantifies the actual contribution of the structural prior.

\begin{table}[htbp]
    \centering
    \caption{Ablation Study (A): Ranking Errors of K-means vs.\ Equal Partition}
    \label{tab:ab}
    \resizebox{\textwidth}{!}{%
    \begin{tabular}{llccccccccc}
        \toprule
        \multirow{2}{*}{\textbf{Dataset}} & \multirow{2}{*}{\textbf{Strategy}} & \multicolumn{9}{c}{\textbf{Labels Per Class ($n_{\mathrm{shot}}$)}} \\
        \cmidrule{3-11}
        & & \textbf{4} & \textbf{6} & \textbf{8} & \textbf{10} & \textbf{12} & \textbf{14} & \textbf{16} & \textbf{18} & \textbf{20} \\
        \midrule
        \multirow{2}{*}{\textbf{Mini-val}}
        & K-means & \textbf{2.87} & \textbf{2.62} & \textbf{2.09} & \textbf{2.00} & \textbf{1.92} & \textbf{1.83} & \textbf{1.70} & \textbf{1.54} & \textbf{1.38} \\
        & Equal Partition & 35.97 & 35.97 & 35.97 & 35.97 & 35.97 & 35.97 & 35.97 & 35.97 & 35.97 \\
        \midrule
        \multirow{2}{*}{\textbf{STL-10}}
        & K-means & \textbf{2.14} & \textbf{1.71} & \textbf{1.60} & \textbf{1.37} & \textbf{1.30} & \textbf{1.29} & \textbf{1.28} & \textbf{1.11} & \textbf{1.18} \\
        & Equal Partition & 13.63 & 14.95 & 15.41 & 15.16 & 17.85 & 13.06 & 16.41 & 14.06 & 17.85 \\
        \midrule
        \multirow{2}{*}{\textbf{CIFAR-10}}
        & K-means & \textbf{3.66} & \textbf{3.48} & \textbf{3.39} & \textbf{3.11} & \textbf{2.83} & \textbf{2.28} & \textbf{2.75} & \textbf{2.40} & \textbf{2.15} \\
        & Equal Partition & 20.87 & 17.28 & 21.89 & 18.91 & 23.25 & 21.89 & 18.99 & 13.91 & 15.89 \\
        \midrule
        \multirow{2}{*}{\textbf{EuroSAT}}
        & K-means & \textbf{5.86} & \textbf{5.13} & \textbf{5.27} & \textbf{5.11} & \textbf{4.53} & \textbf{4.69} & \textbf{4.90} & \textbf{4.66} & \textbf{4.45} \\
        & Equal Partition & 24.56 & 22.33 & 25.42 & 27.25 & 25.59 & 25.86 & 27.95 & 25.39 & 26.54 \\
        \midrule
        \multirow{2}{*}{\textbf{CIFAR-100}}
        & K-means & \textbf{0.95} & \textbf{0.83} & \textbf{0.62} & \textbf{0.51} & \textbf{0.52} & \textbf{0.57} & \textbf{0.37} & \textbf{0.33} & \textbf{0.34} \\
        & Equal Partition & 29.00 & 26.00 & 13.00 & 23.00 & 19.00 & 28.00 & 20.00 & 19.00 & 23.00 \\
        \bottomrule
    \end{tabular}
    }
\end{table}

Table~\ref{tab:ab} shows that K-means consistently yields lower average ranking errors than equal partition across all five datasets and labeling budgets. For example, on CIFAR-100, K-means remains below 1.0 while equal partition ranges from 13.0 to 29.0; on Mini-val, equal partition remains at 35.97, whereas K-means reaches at most 2.87. As the number of labeled samples increases from 4 to 20, K-means generally decreases (e.g., CIFAR-10 from 3.66 to 2.15 and Mini-val from 2.87 to 1.38), although the trend is not strictly monotonic; on STL-10, for instance, it rises from 1.11 to 1.18 at the final budget. These results support the importance of geometry-aware partitioning for EGiME.

\textbf{B. Entropy.} To further investigate the respective contributions of the two entropy terms, we decompose the EGiME formulation $E_r = E_r^{\mathrm{(inter)}} / (1 + E_r^{\mathrm{(intra)}})$ into two ablated variants. Variant~A (inter-cluster entropy only) discards the intra-cluster term by setting the denominator to 1, i.e., $E_r^{\mathrm{A}} = E_r^{\mathrm{(inter)}}$, evaluating models purely by the uniformity of sample dispersion across clusters. Variant~B (intra-cluster entropy only) discards the inter-cluster term by setting the numerator to 1, i.e., $E_r^{\mathrm{B}} = 1 / (1 + E_r^{\mathrm{(intra)}})$, evaluating models purely by the concentration of labeled samples within clusters. We compute the  predicted ranking errors $RE_{\mathrm{variant}}$ across all five datasets by $E_r^{\mathrm{A}}$ and $E_r^{\mathrm{B}}$, and compare them with the ranking errors $RE_{\mathrm{Full}}$ obtained by the full EGiME by $\Delta RE = RE_{\mathrm{variant}} - RE_{\mathrm{Full}}$, as shown in Table~\ref{tab:ab_entropy}, where smaller $\Delta RE$ indicates better prediction.

\begin{table}[htbp]
    \centering
    \caption{The contribution of $E_r^{\mathrm{A}}$ and $E_r^{\mathrm{B}}$. Increase in Ranking Error ($\Delta RE = RE_{\mathrm{variant}} - RE_{\mathrm{Full}}$) after Removing Each Entropy Component. Smaller $\Delta RE$ indicates the removed component is less critical.}
    \label{tab:ab_entropy}
    \small
    \begin{center}
    \begin{tabular}{lccc}
        \toprule
        \multirow{2}{*}{\textbf{Dataset}} & \multicolumn{2}{c}{Average $\Delta RE$} \\
        \cmidrule{2-3}
                         & $E_r^{\mathrm{A}} = E_r^{\mathrm{(inter)}}$  & $E_r^{\mathrm{B}}= 1 / (1 + E_r^{\mathrm{(intra)}})$  \\
        \midrule
        Mini-val  & +1.00 & +2.00  \\
        STL-10    & +1.33 & +1.33 \\
        CIFAR-10  & +3.89 & +3.00 \\
        EuroSAT   & +1.78 & +10.11  \\
        CIFAR-100 & +0.11 & +4.11  \\
        \bottomrule
    \end{tabular}
    \end{center}
\end{table}

Table~\ref{tab:ab_entropy} quantifies the increase in ranking error after removing each entropy component. The inter-cluster term is more influential on three of the five datasets, with the largest effect on EuroSAT ($\Delta RE = +10.11$). CIFAR-10 is an exception: removing the intra-cluster term causes a larger increase ($+3.89$) than removing the inter-cluster term ($+3.00$). On STL-10, the two penalties are equal ($+1.33$). These results indicate that both entropy terms contribute, with a dataset-dependent balance, rather than that one term dominates universally.

\section{Conclusion}
This paper addressed the challenge of evaluating pre-trained neural networks in few-shot transfer learning scenarios, where traditional supervised evaluation methods suffer from severe variance due to scarce labeled data. We proposed EGiME, a semi-supervised evaluation framework based on information-theoretic criteria that leverages both limited labeled samples and abundant unlabeled data. By reformulating model evaluation as the characterization of clustering structure in feature space, EGiME combines inter-cluster entropy, which measures global class separability, with inverse-frequency and intra-cluster entropy, which capture local compactness of labeled anchors. This design yields a training-free, parameter-light, and computationally efficient metric that scales favorably to large model zoos without requiring source-domain data or auxiliary classifier training.

Experiments on five few-shot benchmarks demonstrated that the weighted entropy score correlates strongly with the ground-truth accuracy, obtained by first computing the 10-fold cross-validation accuracy of K-means, KNN, and SVM separately and then averaging the three accuracies with equal weights. EGiME substantially outperforms cross-validation competitors in ranking stability, particularly below 10\% labeling. Ablations confirmed that the adaptive K-means partition is critical, whereas naive equal partitioning severely degrades ranking quality, establishing EGiME as a practical tool for pre-trained model selection in label-scarce regimes.

Several limitations motivate future work: EGiME assumes a known number of target classes and that each class forms a single compact cluster, and the entropy formulation implicitly assumes balanced class distributions. K-means may also perform poorly in very high-dimensional spaces. Addressing these issues, extending the framework to multimodal settings, and investigating its theoretical connections to generalization bounds constitute promising directions.

\section*{Funding}
This work was supported in part by the National Natural Science Foundation of China under Grant 12671626, 12531019, the National Key R \& D Program of China under Grant 2021YFA1003004, and the special foundation for Guangxi Ba Gui Scholars.

\section*{Data availability}
The benchmark datasets used in this study (Mini-val, STL-10, CIFAR-10, EuroSAT, and CIFAR-100) are publicly available. The evaluation code and the per-run ranking results generated during this study are available from the corresponding author on reasonable request.

\section*{Declaration of competing interest}
The authors declare that they have no known competing financial interests or personal relationships that could have appeared to influence the work reported in this paper.

\section*{Author Biographies}

\textbf{Dequan Jin} received the B.Sc., M.Sc., and Ph.D. degrees in mathematics from Xi'an Jiaotong University in 2005, 2008, and 2011. He is currently a Professor with the School of Mathematics and Information Science, Guangxi University. His research interests include mathematical theory, cognition learning, and data analysis.

\textbf{Tao Li} is a graduate student at the School of Mathematics and Center for Applied Mathematics, Guangxi University, Nanning, China. His research focuses on model evaluation in small-sample scenarios.

\textbf{Lingfeng Xing} is a graduate student at the School of Mathematics and Center for Applied Mathematics, Guangxi University, Nanning, China. His research interests lie in 3D Gaussian Splatting and sparse-view X-ray computed tomography reconstruction.

\textbf{Shihui Ying} received the B.Eng. degree in mechanical engineering and the Ph.D. degree in applied mathematics from Xi'an Jiaotong University in 2001 and 2008. He is currently a Professor at Shanghai University. His research interests include machine learning and medical image analysis.

\textbf{Hui Zhao} received the Ph.D. degree in finance from Guangxi University in 2017. She was a visiting scholar at the University of Hong Kong and Fordham University. She is currently a Professor with the School of Economics, Guangxi University. Her research interests include data elements, data governance, and China-ASEAN economic cooperation.

\bibliographystyle{elsarticle-num}
\bibliography{EGIME_refs}

\end{document}